\documentclass[conference]{IEEEtran}
\IEEEoverridecommandlockouts
\usepackage{cite}
\usepackage{amsmath,amssymb,amsfonts}
\usepackage{algorithmic}
\usepackage{graphicx}
\usepackage{subfigure}
\usepackage{textcomp}
\usepackage{xcolor}
\usepackage[linesnumbered,ruled,vlined]{algorithm2e}
\usepackage{cuted}

\usepackage{titlesec}
\usepackage{hyperref}

\titleclass{\subsubsubsection}{straight}[\subsection]

\newcounter{subsubsubsection}[subsubsection]
\renewcommand\thesubsubsubsection{\thesubsubsection.\arabic{subsubsubsection}}

\titleformat{\subsubsubsection}
  {\normalfont\normalsize\bfseries}{\thesubsubsubsection}{1em}{}
\titlespacing*{\subsubsubsection}
{0pt}{3.25ex plus 1ex minus .2ex}{1.5ex plus .2ex}

\makeatletter
\renewcommand\paragraph
  {\@startsection{paragraph}{5}{\z@}%
  {3.25ex \@plus1ex \@minus.2ex}%
  {-1em}%
  {\normalfont\normalsize\bfseries}}
  
\renewcommand\subparagraph{\@startsection{subparagraph}{6}{\parindent}%
  {3.25ex \@plus1ex \@minus .2ex}%
  {-1em}%
  {\normalfont\normalsize\bfseries}}
\def\toclevel@subsubsubsection{4}
\def\toclevel@paragraph{5}
\def\toclevel@paragraph{6}
\def\l@subsubsubsection{\@dottedtocline{4}{7em}{4em}}
\def\l@paragraph{\@dottedtocline{5}{10em}{5em}}
\def\l@subparagraph{\@dottedtocline{6}{14em}{6em}}
\makeatother

\def\BibTeX{{\rm B\kern-.05em{\sc i\kern-.025em b}\kern-.08em
    T\kern-.1667em\lower.7ex\hbox{E}\kern-.125emX}}

\begin{document}

\title{A Selective Survey on Versatile Knowledge Distillation Paradigm for Neural Network Models
}

\author{Jeong-Hoe Ku, JiHun Oh, YoungYoon Lee, Gaurav Pooniwala, SangJeong Lee \\
\IEEEauthorblockA{\textit{Samsung Research}, \\
\textit{Samsung Electronics Co., Ltd.}\\
Seoul, Republic of Korea \\
Email: \{mrku69, jihun2331.oh, euler.lee, pt.gaurav, sj94.lee\}@samsung.com
}
}

\maketitle
\thispagestyle{plain}
\pagestyle{plain}

\begin{abstract}
This paper aims to provide a selective survey about \emph{knowledge distillation}~(KD) framework for researchers and practitioners to take advantage of it for developing new optimized models in the deep neural network field. To this end, we give a brief overview of knowledge distillation and some related works including \emph{learning using privileged information}~(LUPI) and \emph{generalized distillation}~(GD). Even though knowledge distillation based on the teacher-student architecture was initially devised as a model compression technique, it has found versatile applications over various frameworks. In this paper, we review the characteristics of knowledge distillation from the hypothesis that the three important ingredients of knowledge distillation are \emph{distilled knowledge and loss}, \emph{teacher-student paradigm}, and the \emph{distillation process}. In addition, we survey the versatility of the knowledge distillation by studying its direct \emph{applications} and its usage in combination with other \emph{deep learning paradigms}. Finally we present some future works in knowledge distillation including explainable knowledge distillation where the analytical analysis of the performance gain is studied and the self-supervised learning which is a hot research topic in deep learning community. 
\end{abstract}

\section{Introduction}

Deep neural networks have been applied to various tasks and achieved dramatic success in many fields. However, larger neural networks with more layers and nodes are used to achieve higher performance. Many recent successful deep neural models are computationally expensive and memory intensive~\cite{Goodfellow2016}. Since it is difficult to deploy such heavy models on devices with low system resources for running real-time applications with a strict latency requirement, it is crucial to reduce their parameter size and computational complexity without performance degradation.

Yu Cheng {\it et al}.~\cite{Cheng2019} briefly categorized compression techniques for the purpose into four schemes: \emph{parameter pruning and sharing, low-rank factorization, transferred/compact convolution filters, quantization}, and \emph{knowledge distillation}, each of which has it's own advantages and drawbacks. Among these techniques, we focus on {\bf\emph{knowledge distillation}} as it is an empirically very successful technique for knowledge transfer between classifiers in an interactive manner which is more similar to how humans learn. Knowledge distillation with neural networks was pioneered by Hinton {\it et al.}~\cite{Hinton2015} which is a transfer learning method that aims to improve the training of a student network by relying on knowledge borrowed from a powerful teacher network.

This paper is organized as follows: \\
In Section~\ref{sec:KD}, we explain knowledge distillation framework briefly and reviewed three important design factors of knowledge distillation including Distilled Knowledge and Loss, the Teacher-Student architecture and the Distillation Process to investigate the internals of knowledge distillation framework. 
In Section~\ref{sec:application}, we survey versatile knowledge distillation framework usages in the fields of Computer Vision, Natural Language Processing and Quantization. 
In Section~\ref{ssec:DL}, we look at the position of Knowledge distillation in the broader Deep learning paradigms of Supervised Learning, Weakly-Supervised Learning and Semi-Supervised/Unsupervised Learning. 
In Section~\ref{sec:related} we look at a few techniques that do not fit the definition of Knowledge distillation but use very similar ideas.
In Section~\ref{sec:future}, we suggest two topics as future works of knowledge distillation

\begin{figure*}[htbp]
\centering
\includegraphics[width=1.0\linewidth]{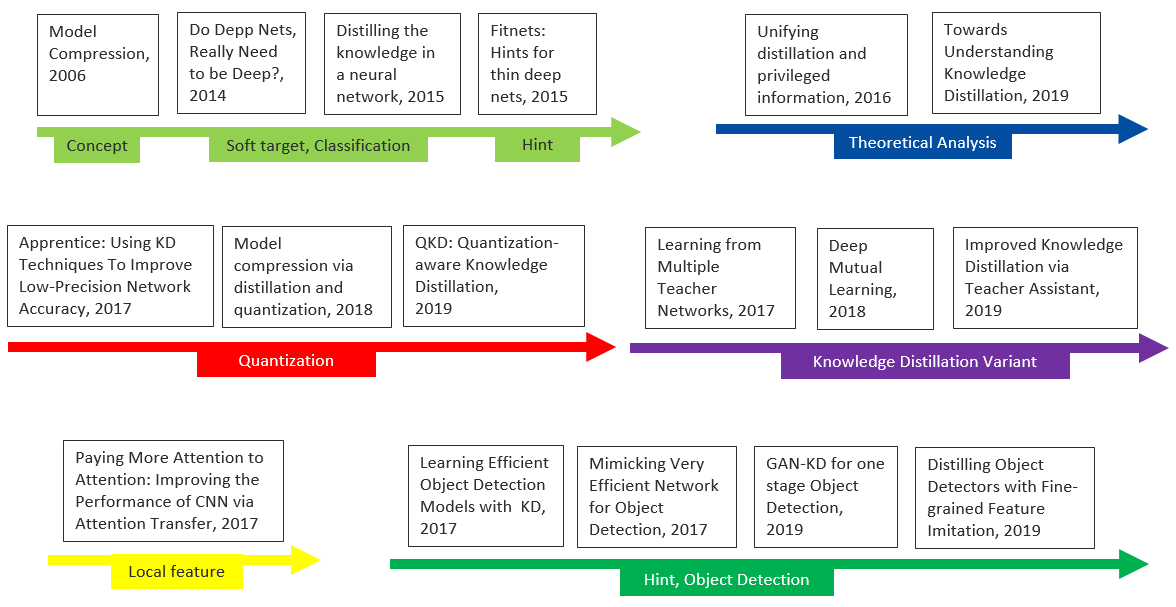}

\caption{{\bf{Overall Knowledge Distillation (KD) Scope.}} Overall Knowledge Distillation (KD) Scope. Overall scope of Knowledge Distillation (KD) and its related technical domains surveyed in this paper. Each arrow groups some related topics (represented as the title of the corresponding paper) chronically in that domain.}
\label{fig:Annex_1}
\end{figure*}

\begin{figure*}[htbp]
\centering
\includegraphics[width=1.0\linewidth]{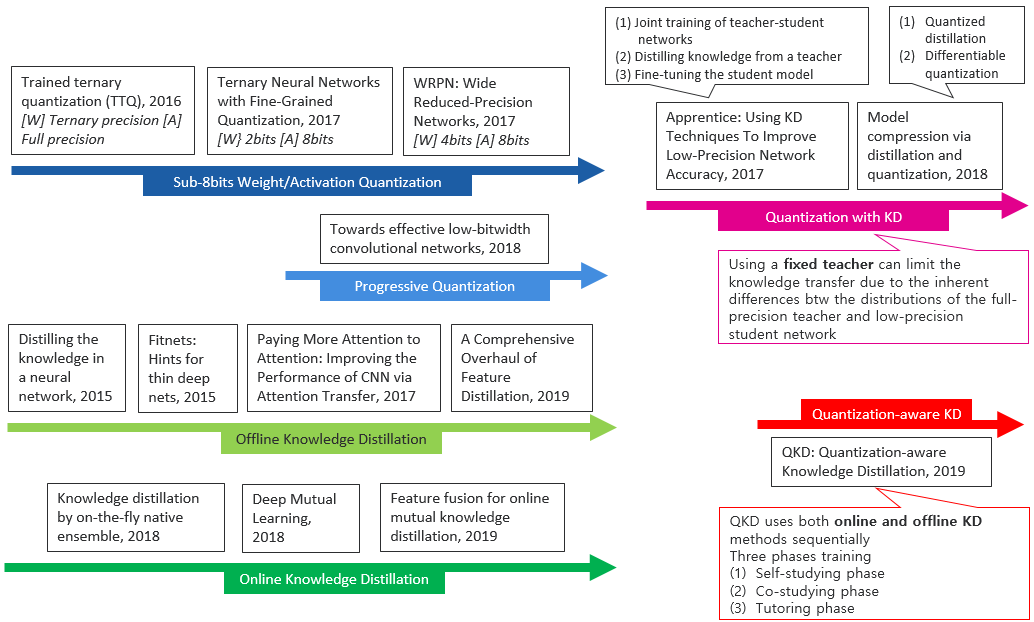}

\caption{{\bf{Knowledge Distillation Applied to Quantization.}} Overallscope where knowledge distillation applied to quantization}
\label{fig:Annex_2}
\end{figure*}

\begin{figure*}[htbp]
\centering
\includegraphics[width=1.0\linewidth]{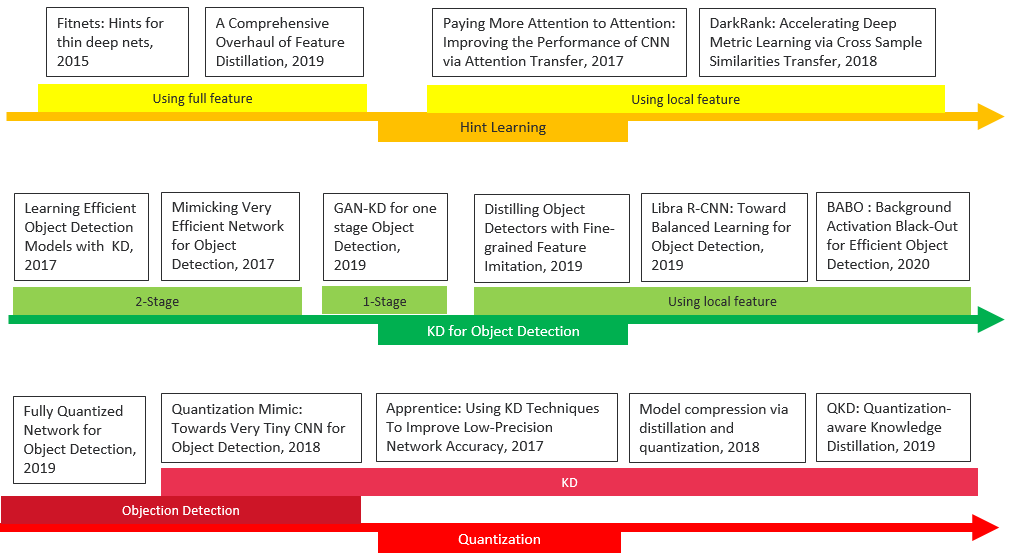}

\caption{{\bf{Knowledge Distillation Applied to Object Detection.}} Overall scope where knowledge distillation is applied to object detection}
\label{fig:Annex_3}
\end{figure*}

\begin{figure*}[htbp]
\centering
\includegraphics[width=1.0\linewidth]{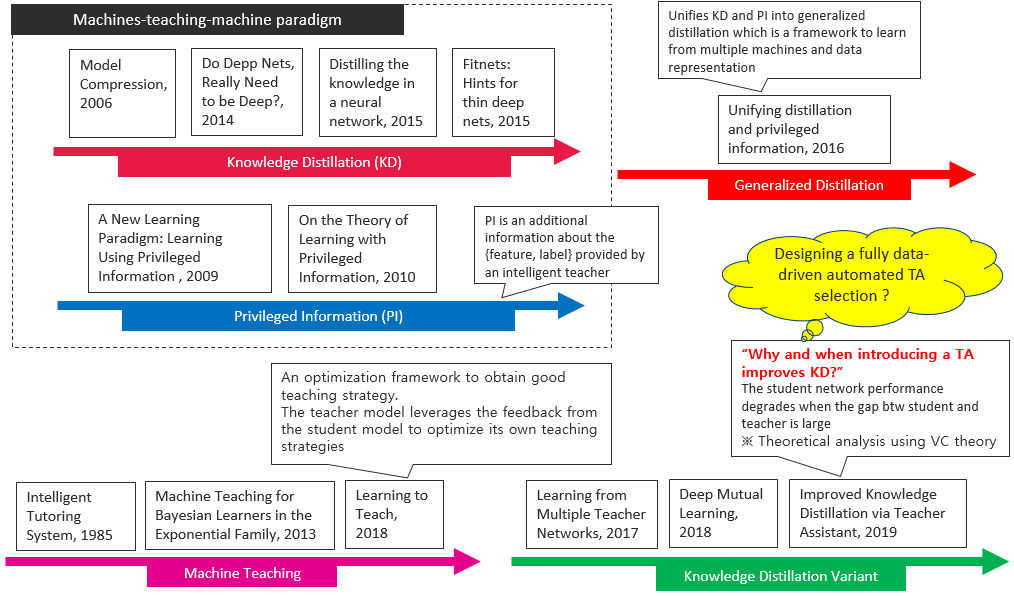}

\caption{{\bf{Knowledge Distillation Variants, Learning Using Privileged Information (LUPI).}} Overall scope which represents the relationship among knowledge distillation, learning with privileged information and generalized distillation}
\label{fig:Annex_4}
\end{figure*}

Fig.~\ref{fig:Annex_1}, Fig.~\ref{fig:Annex_2}, Fig.~\ref{fig:Annex_3}, and Fig.~\ref{fig:Annex_4} in Annex show the brief overview of the topics of knowledge distillation reviewed in this survey paper.

\section{Knowledge Distillation} \label{sec:KD}
Knowledge distillation is an effective model compression technique in which a compact model (student) is trained under the supervision of a larger pre-trained model or an ensemble of models (teacher).

Knowledge distillation aims to improve the performance of the student network by providing additional supervision from a teacher network. 
To the best of our knowledge, exploiting knowledge transfer to compress model was first proposed in C. Bucilu\v{a} {\it et al.}~\cite{Bucilua2006}. They trained a compressed/ensemble model of strong classifiers with pseudo-labeled data, and reproduced the output of the original larger network. However, the work is limited to shallow models. The idea has been adopted in~\cite{Ba2014} as knowledge distillation to compress deep and wide networks into shallower ones, where the compressed model mimicked the function learned by the complex model. Hinton {\it et al.}~\cite{Hinton2015} popularized the concept of Knowledge Distillation to be extended to more practical uses. The work in~\cite{Hinton2015} proposed knowledge distillation as a more general case of C. Bucilu\v{a} {\it et al.}~\cite{Bucilua2006} by adopting the concept of temperature parameter at the output of teacher. The student was trained to predict the output and the classification labels. 

The main idea of knowledge distillation approach is to shift knowledge from a large teacher model into a small one by learning the class distributions output via softmax~\cite{Hinton2015}. It has even been observed that the student learns much faster and more reliably if trained using outputs of teacher as soft labels, instead of one-hot-encoded labels.

Since then, a number of knowledge distillation methods have been proposed, each trying to capture and transfer some characteristics of the teacher such as the representation space, decision boundary or intra-data relationship. Despite its simplicity, knowledge distillation demonstrates promising results in various image classification tasks.

Knowledge distillation has proven empirically to be an effective technique for training a compact model~\cite{Romero2015, Kim2019, Mirzadeh2019}.

\subsection{Distilled Knowledge and Loss} \label{ssec:knowloss}
Despite the recent advances of knowledge distillation technique, a clear understanding of where knowledge resides in a deep neural network and an optimal method for capturing knowledge from teacher and transferring it to student remains an open question. 

In recent advances of knowledge distillation, many forms of knowledge have been defined (Jiaxi Tang {\it et al.}, 2020~\cite{Tang2020}) based on the teacher-student learning paradigm and have shown dramatic success and were analyzed empirically:
\begin{itemize}
\item Layer activation~\cite{Romero2015}
\item Auxiliary information~\cite{Vapnik2015}
\item Jacobian matrix of the model parameters~\cite{Czarnecki2017, Srinivas2018}
\item Gram matrix derived from pairs of layers~\cite{Yim2017}
\item Activation boundary~\cite{Heo2019}
\end{itemize}

Distillation loss for knowledge distillation training is a key factor which is used to penalize the student to transfer this Knowledge from the Teacher to the Student. 

Fahad Sarfraz {\it et al.}~\cite{Sarfraz2020} presented broad categorization of a diverse set of knowledge distillation methods which differ from each other with respect to how knowledge is defined and transferred from the teacher. Borrowing from their categorization, we cite two groups below to demonstrate how to capture the knowledge from teacher.

\textbf{a) Response Distillation} uses only the outputs of a Teacher to train the student to mimic it.
C. Bucilu\v{a} {\it et al.}~\cite{Bucilua2006} proposed to use the logits of a teacher network as target for the student and to minimize the squared difference. 
Hinton {\it et al.}~\cite{Hinton2015} proposed to minimize the KL divergence between the smoother output probabilities. 
In the original formulation, Hinton {\it et al.}~\cite{Hinton2015} introduced a knowledge distillation compression framework and proposed mimicking the softened softmax output of the teacher using a temperature parameter. It raised the temperature of the final softmax function and minimize the Kullback-Leibler (KL) divergence between the smoother output probabilities. This softened output transfers more important information which is called {\it dark knowledge} compared to the hard output. When the soft targets have high entropy, they provide much more information per training case than hard targets and much less variance in the gradient between training cases, so the small model can often be trained on much less data than the original cumbersome model while using a much higher learning rate. 

When the dimension of both outputs are the same, these methods can be applied to any pair of network architectures. Although this loss was originally proposed to apply to a simple task such as image classification, it has seen a wide variety of applications.

\textbf{b) Representation Space Distillation} aims to mimic the latent feature space of the teacher. Adriana Romero {\it et al.}~\cite{Romero2015} introduced intermediate level hints from the teacher’s hidden layers to guide the training process of the student. Due to the differences of a dimension of hidden layers between the teacher and student network, design of the feature distillation method needs to be done carefully to prevent information loss when transferring knowledge. 
Byeong Heo {\it et al.}~\cite{Heo2019} proposed a novel feature distillation method and designed a new distillation loss to minimize the information loss. This gave a hint to extend knowledge distillation to more complex tasks such as object detection.

\subsection{Teacher-Student Architecture} \label{ssec:TSarch}
Knowledge distillation has proven to be an effective technique for training a compact model and also providing greater architectural flexibility since it allows for structural differences in the teacher and student networks. There are several variants of the conventional knowledge distillation Student and Teacher architectures which extend the student-teacher learning paradigm to improve the performance and overcome some weak points. 

\subsubsection{Single Teacher- Single Student} \label{ssec:single}
In the perspective of a Single Teacher- Single Student learning paradigm, knowledge distillation is a simple way to transfer knowledge of a teacher to improve the performance of small deep learning model called a student. More specifically, knowledge distillation refers to the method that helps the training process of a small network (student) under the supervision of a large network (teacher). The additional supervision about the relative probabilities of secondary class and relational information between data points at the output of Teacher network can be useful in increasing the efficacy of the Student network.

We can downsize a student network regardless of the structural difference between teacher and student. 
Allowing this architectural flexibility, knowledge distillation is emerging as a next generation approach of network compression. However, too excessive gap of the capacity between a teacher network and a student network is a critical obstacle for knowledge transfer performance. This is empirically shown and analyzed in~\cite{Mirzadeh2019}.

\subsubsection{Multi-Step Learning}
Seyed-Iman Mirzadeh {\it et al.}~\cite{Mirzadeh2019} showed that the student network performance degrades when the gap between student and teacher networks is large. So, given a fixed student network, one cannot employ an arbitrarily large teacher network; in other words, a teacher network can effectively transfer its knowledge to student networks having up to a certain capacity. 

To alleviate this shortcoming, multi-step knowledge distillation was introduced. It is a new distillation framework called \emph{Teacher Assistant Knowledge Distillation}~(TAKD), which introduces intermediate-sized network known as teacher assistants (TAs) between the teacher and the student to fill in the gap. TA models are distilled from the teacher, and the student is then only distilled from the TAs. Through extensive empirical evaluations and a theoretical justification, they showed that introducing intermediate TA networks improve the distillation performance and concluded that the size (capacity) gap difference between a teacher-TA and a TA-student is important.

\begin{figure}[htbp]
\centering
\centerline{\includegraphics[width=\linewidth]{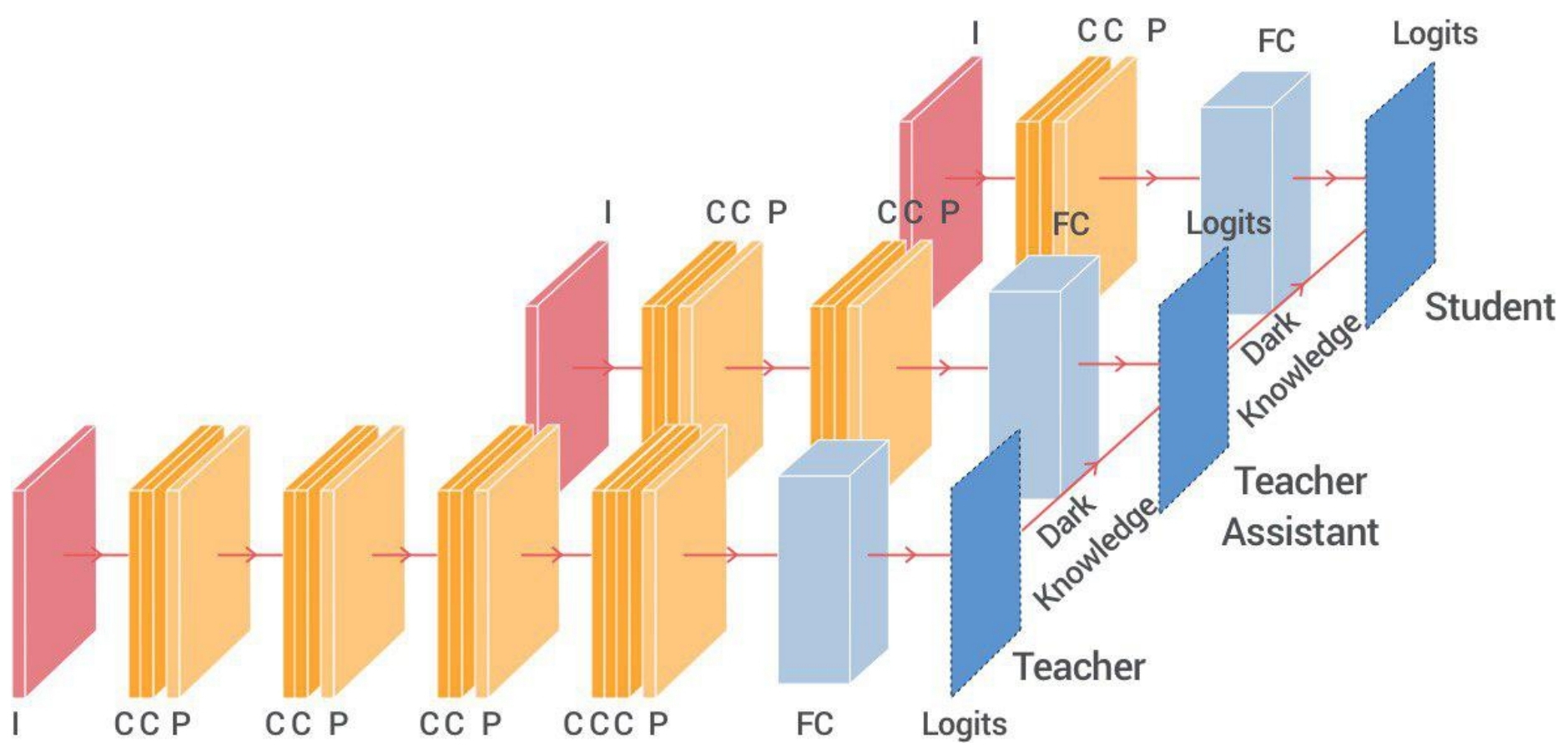}}
\caption{A teacher assistant network fills the gap between student and teacher networks~\cite{Mirzadeh2019}.}
\label{fig:Mirzadeh2019}
\end{figure}

Fig.~\ref{fig:Mirzadeh2019} shows the overall knowledge distillation structure incorporating teacher assistant.

\subsubsection{Multiple-Teacher Learning} \label{ssec:multiteach}
\begin{figure}[htbp]
\centerline{\includegraphics[width=\linewidth]{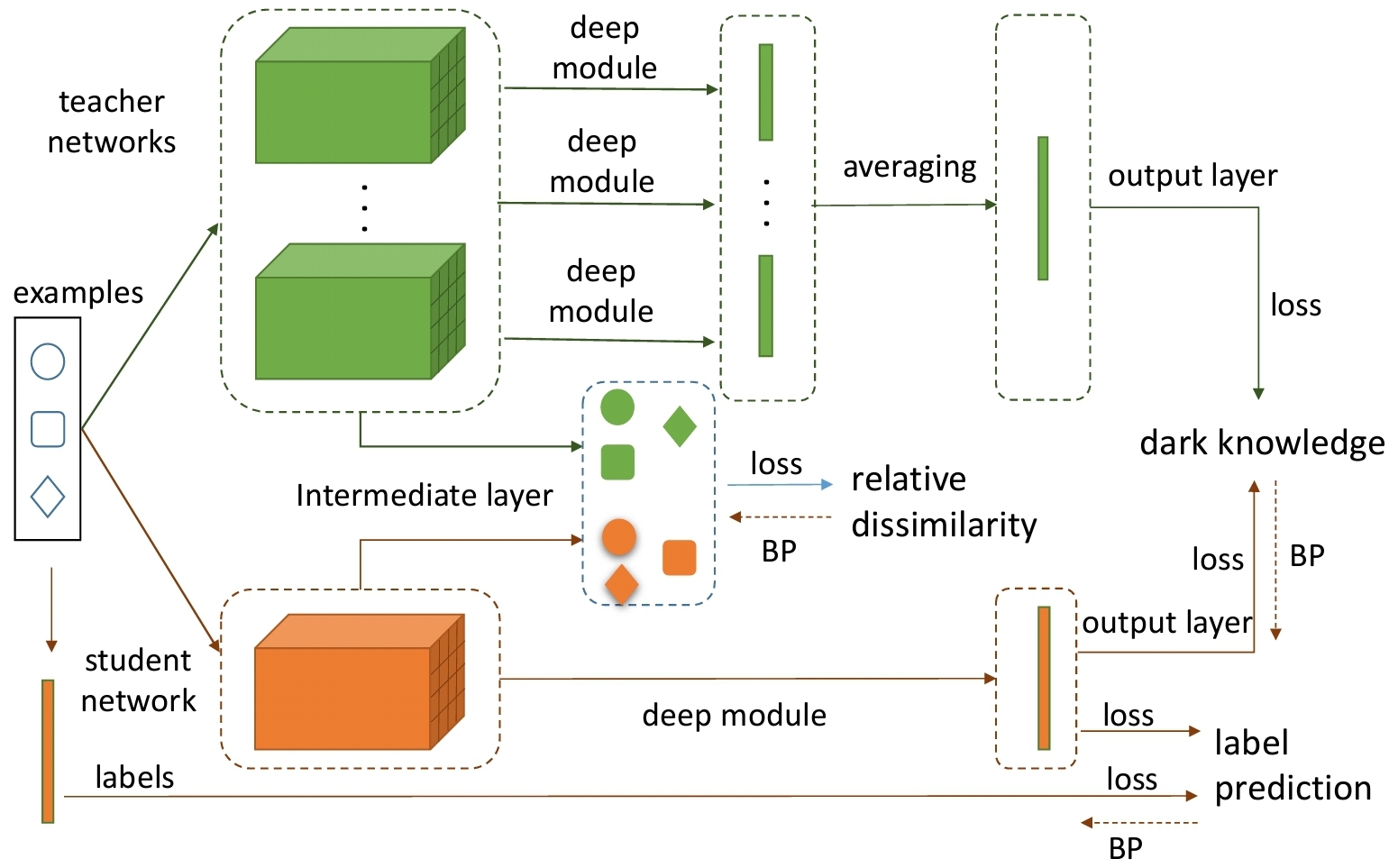}}
\caption{A graphical diagram for the proposed method to train a new thin deep student network by incorporating multiple comparable teacher networks~\cite{You2017}.}
\label{fig:You2017}
\end{figure}

Shan You {\it et al.}~\cite{You2017} proposed a new method which uses  multiple teacher networks to train a thin and deep student network. 
Fig.~\ref{fig:You2017} shows overall knowledge distillation incorporating multiple teachers.
The method uses three losses to train the Student: a label prediction loss, a dark knowledge loss and a relative similarity loss. The incorporation of multiple teacher networks exists in two places. 

One is in the output layers via averaging the softened output targets(dark knowledge) from different Teacher networks and using it to train the Student. 

It also used knowledge about the intermediate layers by imposing a constraint of the dissimilarity among examples. The authors suggest that relative dissimilarity between intermediate representations of different examples serves as more flexible and appropriate guidance from teacher networks.

\subsection{Distillation Process} \label{ssec:process}

\subsubsection{Off-line Distillation}
In vanilla knowledge distillation (Hinton et al.,~\cite{Hinton2015}), we start with a powerful large and pre-trained teacher network and perform one-way knowledge transfer to a small untrained student. This is known as offline knowledge distillation. Most previous Knowledge distillation methods use this process. The large teacher model is first trained on a set of training samples. The teacher model is then used to extract the knowledge in the forms of logits or the intermediate features, which are then used to guide the training of the student model during distillation.

The teacher models typically need to have a high capacity and require a lot of time and data for training. The training of the student model in offline distillation is usually efficient under the guidance of the teacher model. A capacity gap between large teacher network and small student network always exists and Seyed-Iman Mirzadeh {\it et al.}~\cite{Mirzadeh2019} showed that the student network performance degrades when the gap is too large.

\subsubsection{On-line Distillation} \label{mutual}

\begin{figure}[htbp]
\centerline{\includegraphics[width=\linewidth]{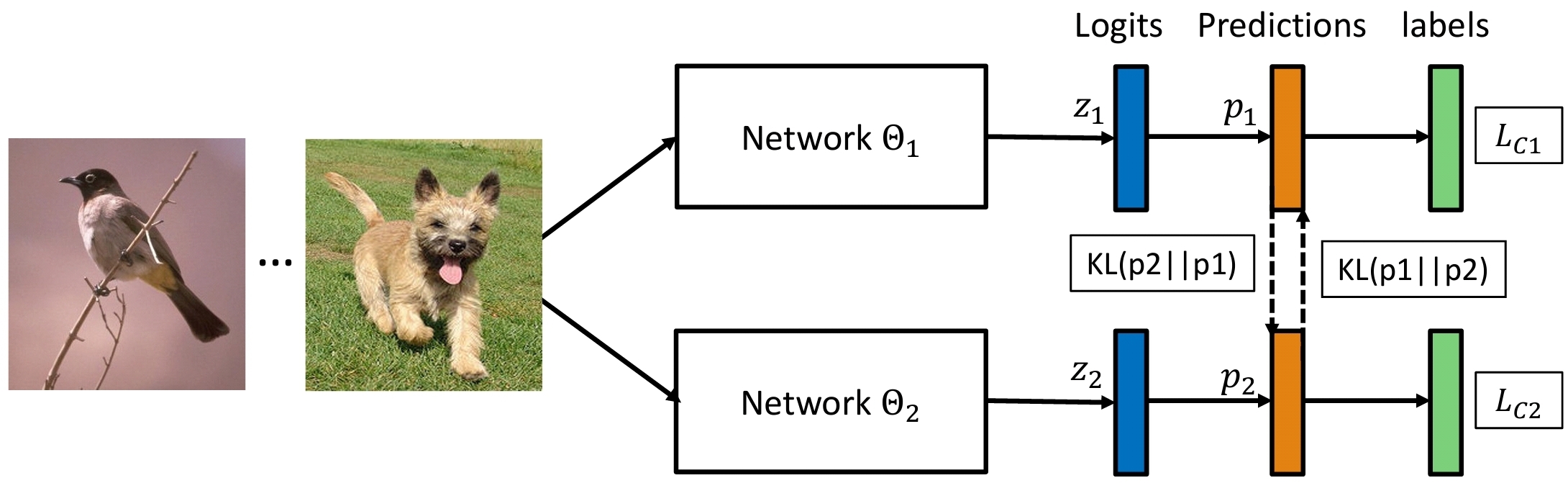}}
\caption{The Deep Mutual Learning (DML) schematic~\cite{Zhang2017}.}
\label{fig:Zhang2017}
\end{figure}
Online distillation is proposed to further improve the performance of the student
model, especially when a large-capacity high performance teacher model is not available (Zhang et al.~\cite{Zhang2017}, ; Chen et al., 2020c). In online distillation, both
the teacher model and the student model are updated simultaneously, and the whole knowledge distillation framework is end-to-end trainable

Ying Zhang et al.~\cite{Zhang2017} first presented an online distillation strategy - deep mutual learning (DML). Rather than one way transfer between a static pre-defined teacher network and a student network, an ensemble of student networks learns collaboratively and teaches each other throughout the training process. 
DML starts with a pool of untrained students who learn simultaneously to solve the task together. Specifically, each student is trained using two losses: a conventional supervised learning loss and a mimicry loss that aligns each student’s class posterior with the class probabilities of other students. Fig.~\ref{fig:Zhang2017} shows the deep mutual learning method.
Each network uses for training a supervised learning loss and a KLD-based mimicry loss to match the probability estimates of its peers.

Guo {\it et al.}~\cite{Guo2020} improved the generalization ability of this procedure by using an ensemble of soft logits. Chen {\it et al.} ~\cite{Chen2020c} performed two-level distillation during training with multiple auxiliary peers and one group
leader to form a diverse set of peer models.

Kim {\it et al.}~\cite{Kim2019} used a feature fusion module to construct the teacher classifier.
An ensemble of sub-network classifiers transfers its knowledge to the fused classifier and then the fused classifier delivers its knowledge back to each sub-network, mutually teaching one another in an online-knowledge distillation manner.

\section{KD Applications} \label{sec:application}

Knowledge distillation is a flexible approach that can be leveraged in many conventional techniques and applications. In this paper we selected three topics where knowledge distillation is actively applied.

\subsection{KD in Computer Vision (CV)}
An excellent example of the successes in computer vision related deep learning can be illustrated with the ImageNet Challenge~\cite{Russakovsky2015}. This challenge is a contest involving two different components, image classification and object detection tasks. 

\subsubsection{KD in Image Classification}
Since knowledge distillation was first introduced in \cite{Bucilua2006}, the initial application was the fundamental computer vision task of image classification~\cite{Ba2014,Hinton2015} and demonstrated excellent improvements. The pioneering study on knowledge distillation~\cite{Hinton2015} showed that a shallow or compressed model trained to mimic the behavior of a deeper or more complex model can recover some or all of the accuracy drop

\subsubsection{KD in Object Detection}
Most modern object detection methods make predictions relative to some initial guesses. Two-stage detectors~\cite{SRen2015, Cai2019} predict boxes w.r.t. proposals, whereas single-stage methods make predictions w.r.t. anchors~\cite{Lin2017} or a grid of possible object centers~\cite{Zhou2019, Tian2019}. 

Applying knowledge distillation techniques to multi-class object detection is challenging for several reasons. First, the performance of detection models suffers from more degradation after compression since detection labels are more expensive and thereby usually less voluminous. Second, knowledge distillation is proposed for classification assuming each class is equally important, whereas that is not the case for detection where the background class is far more prevalent. Third, detection is a more complex task that combines elements of both classification and bounding box regression. 

We review four papers~\cite{GuobinChen2017, Wang2020, Li2017, TaoWang2019} where knowledge distillation was applied to object detection.

\paragraph{Global Feature based Object Detection} 

It is difficult to apply vanilla knowledge distillation architecture ~\cite{Hinton2015} to object detection because it only transfers soft-target that is the logit of penultimate layer. It is necessary to transfer more knowledge for object detection task to the student network. Romero {\it et al.}~\cite{Romero2015} first proposed the framework called FitNet where the student network mimics the full feature maps of the teacher network. The intermediate representations of teacher networks are called ‘Hint’. FitNet opened a way to apply knowledge distillation to object detection task and since then researchers proposed many forms of object detection frameworks using knowledge distillation.

\begin{figure}[htbp]
\centerline{\includegraphics[width=\linewidth]{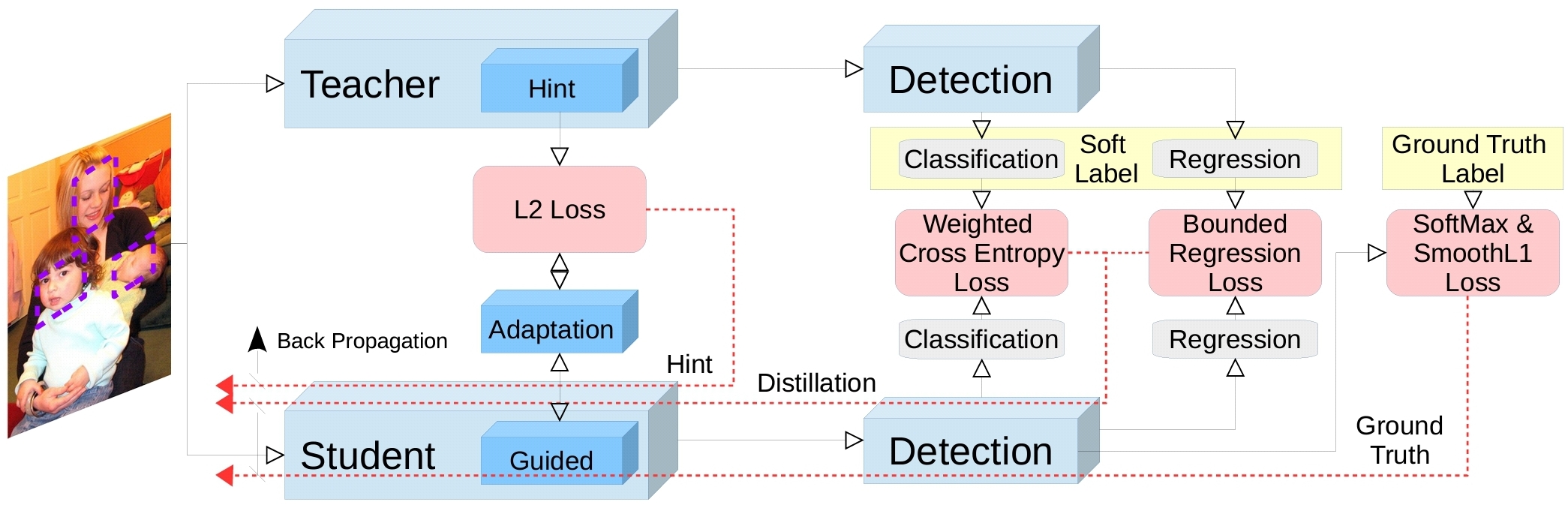}}
\caption{The proposed learning scheme on a visual object detection task using Faster-RCNN, which mainly consists of the region proposal network (RPN) and the region classification network (RCN)~\cite{GuobinChen2017}.}
\label{fig:GChen2017}
\end{figure}
Guobin Chen {\it et al.}~\cite{GuobinChen2017} proposed a new end-to-end trainable framework to train compact and fast multi-class object detection networks with improved accuracy using knowledge distillation~\cite{Hinton2015} and hint learning~\cite{Romero2015}. Two losses are proposed to effectively address the aforementioned challenges. One is a weighted cross entropy loss for classification that accounts for the imbalance in the impact of misclassification for background class as opposed to object classes. The other is a teacher bounded regression loss for knowledge distillation. For hint learning, adaptation layers are provided to allow the student to better learn from the distribution of neurons in intermediate layers of the teacher. 
Fig.~\ref{fig:GChen2017} shows the specialized knowledge distillation scheme proposed for applying to Faster R-CNN. 
The two networks both use the multi-task loss to jointly learn the classifier and the bounding-box regressor. We employ the final output of the teacher model’s RPN and RCN as the distillation targets, and apply the intermediate layer outputs as a hint. Red arrows indicate the backpropagation pathways.

\begin{figure}[htbp]
\centerline{\includegraphics[width=\linewidth]{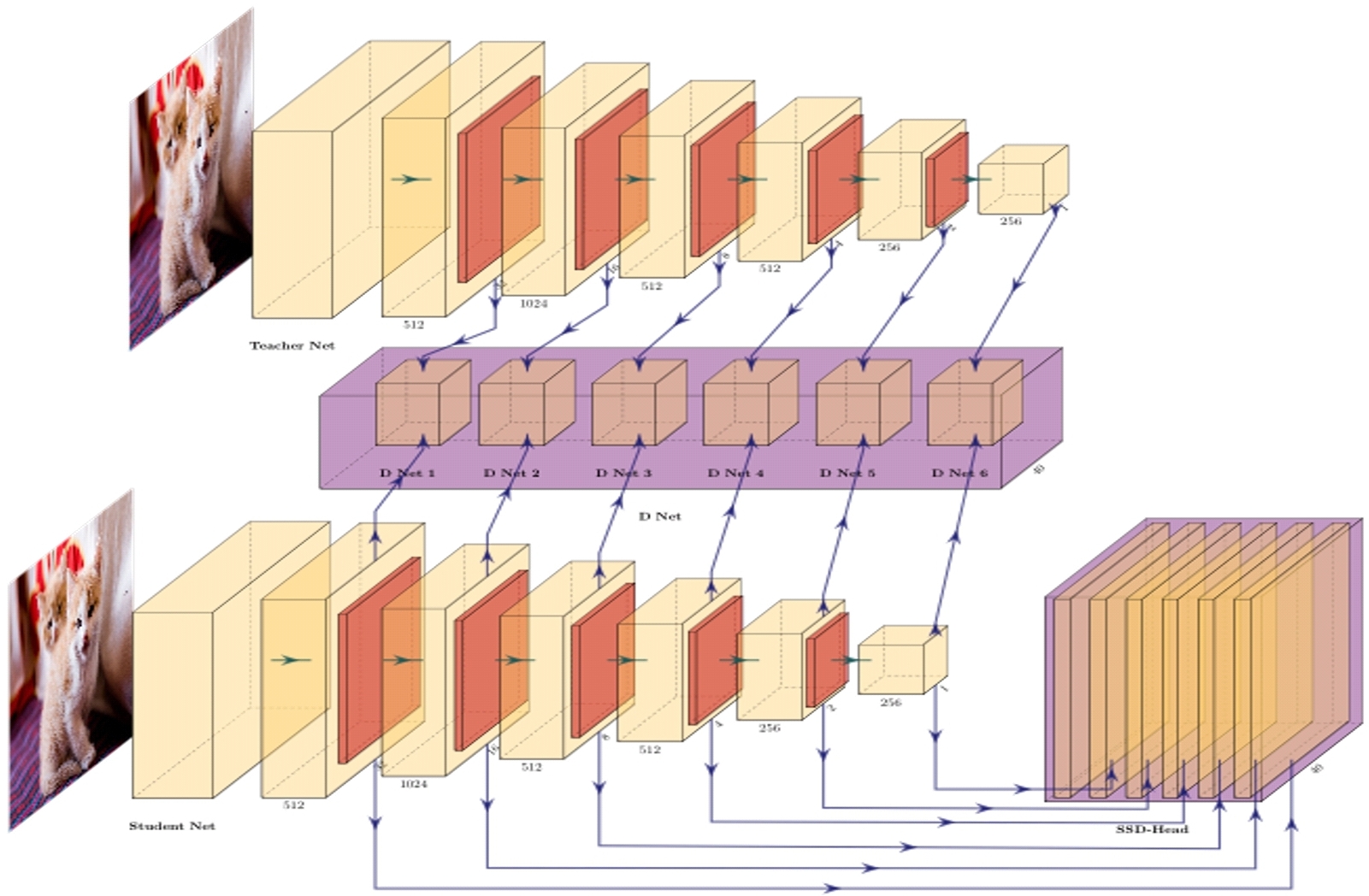}}
\caption{The GAN-KD network architecture~\cite{Wang2020}.}
\label{fig:Wang2020}
\end{figure}
Wanwei Wang {\it et al.}~\cite{Wang2020} proposed a clean and effective knowledge distillation method for single-stage object detection called GAN-KD. The feature maps generated by a teacher network and a student network are used as true samples and fake samples respectively, and it conducts adversarial training for both to improve the performance of the student network in single-stage object detection. The GAN algorithm is employed to complete the migration from a teacher network to a student network. GAN-KD is composed of four modules: a teacher network, a student network, a discriminative network (D-Net), and a SSD-Head network as shown below. D-Net learns to determine whether a sample is from the teacher network or the student network.
Fig.~\ref{fig:Wang2020} shows the GAN-KD network architecture.
Teacher Net (the top left) is a backbone network of the larger and fully trained SSD model. Student Net (the bottom left) is a smaller network such as MobileNet. SSD-Head (the bottom right) is the head network of the SSD model which is responsible for the classification and anchor box regression. D-Net is a module consisting of six small discriminant networks.

Unlike the traditional knowledge distillation algorithm, there is no need to manually specify the location of knowledge distillation. It does not require the design of complex cost functions, and can be applied to one-stage object detection.

\paragraph{Local Feature based Object Detection}

When applying the whole feature map, the object correspondence information might be ignored or degraded especially for small object feature learning. Compared to the global context features, the features of local regions contain more representative information for object detection. Therefore, Byungseok Roh {\it et al.}~\cite{Roh2020} proposed objectness-aware object detection method. There have been some knowledge distillation approaches to leverage local feature based object detection utilizing local features when distilling knowledge from a teacher network to a student network effectively.

Quanquan Li {\it et al.}~\cite{Li2017} presented a feature map mimicking method aiming to train the small model to mimic the feature map activations of the large model in an unified fully convolutional network object detection pipeline. The feature map matters in object detection since both the objectness scores and locations are predicted based on the feature map. Therefore, it is more reasonable to mimic the output feature maps between the two detection networks which contain the response information across an entire image.

\begin{figure}[htbp]
\centerline{\includegraphics[width=\linewidth]{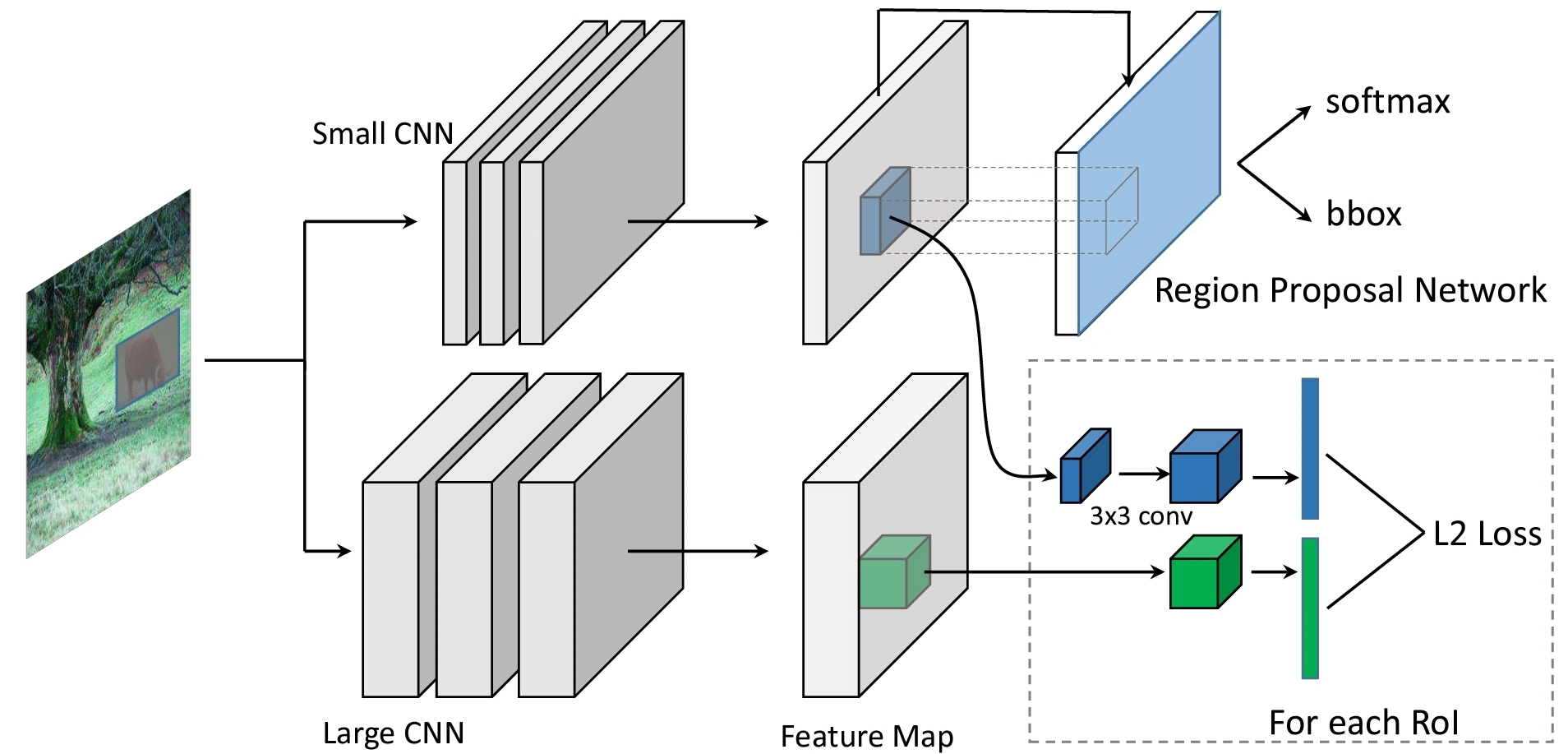}}
\caption{An overall architecture of feature mimicking by proposal sampling~\cite{Li2017}.}
\label{fig:Li2017}
\end{figure}
Fig.~\ref{fig:Li2017} shows the overall architecture based on the RoI-aware knowledge distillation.
A Region Proposal Network generates candidate ROIs, which then are used to extract local features from the feature maps.

The work in~\cite{Li2017} proposed to only transfer knowledge inside the area of proposals. However, the mimicking regions depend on the output structure and internal information of model itself, so it is not applicable to one-stage detector.

\begin{figure}[htbp]
\centerline{\includegraphics[width=\linewidth]{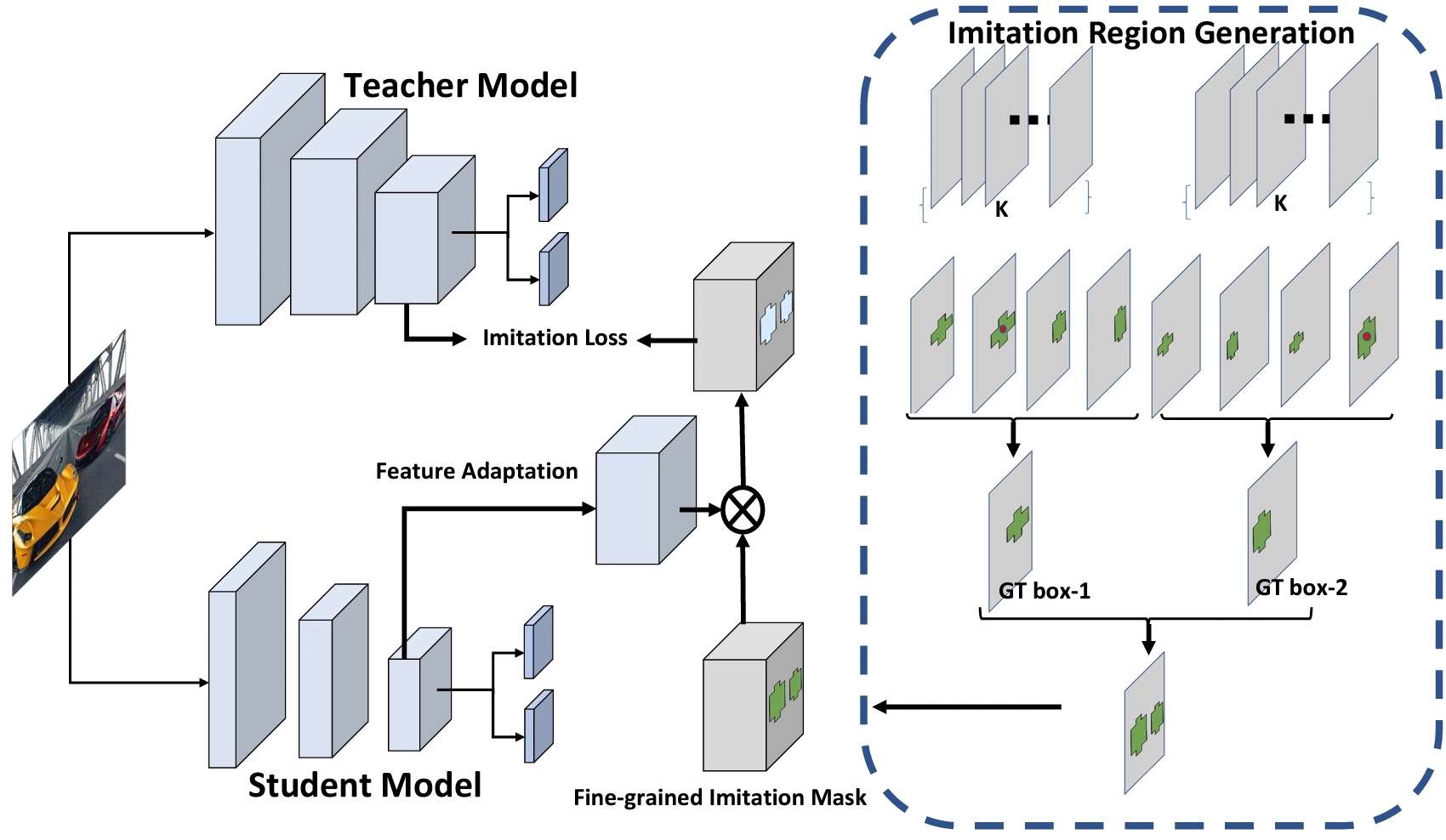}}
\caption{An illustration of the proposed fine-grained feature imitation method~\cite{TaoWang2019}.}
\label{fig:Wang2019}
\end{figure}
Tao Wang {\it et al.}~\cite{TaoWang2019} proposed a fine-grained feature imitation method exploiting the cross-location discrepancy of feature response based on the intuition that detectors have attention to the local regions near the object regions. Thus the discrepancy of feature response on the near object anchor locations reveals important information of how a teacher model learns. The novel mechanism is designed to estimate those locations and let student model imitate the teacher on them to achieve the enhanced accuracy.
Fig.~\ref{fig:Wang2019} shows the proposed fine-grained feature imitation method.
The student detector is trained by both ground-truth supervision and imitation of teacher’s feature response on close object anchor locations. The feature-adaptation layer makes student’s guided feature layer compatible with the teacher. To identify informative locations, it iteratively calculates IOU map of each ground-truth bounding box with anchor priors, filter and combine candidates, and generate the final imitation mask.

Recently, Yongcheng Liu {\it et al.}~\cite{Liu2019} applied ROI-aware distillation approach to the weakly-supervised detection problem.

\subsection{Natural Language Processing} \label{ssec:NLP}

The up-to-date language model, such as BERT, has significantly improved the performance of many natural language processing. However, the pre-trained language models are computationally expensive and memory intensive, so it is difficult to effectively execute them on resource-restricted devices. 

\subsubsection{TinyBERT}
\begin{figure}[htbp]
\centering
\subfigure[ ]{
\includegraphics[width=0.48\linewidth]{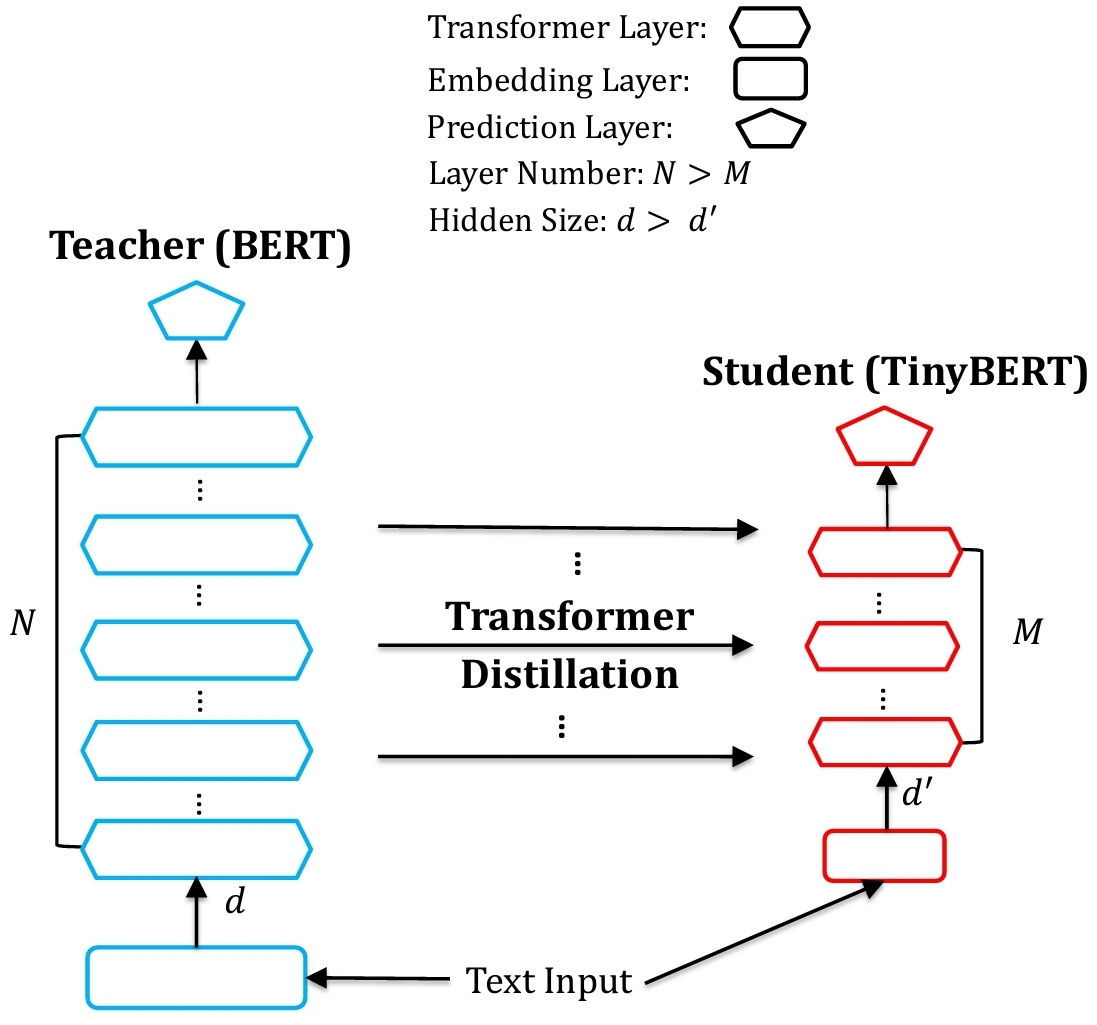}}
\subfigure[ ]{
\includegraphics[width=0.48\linewidth]{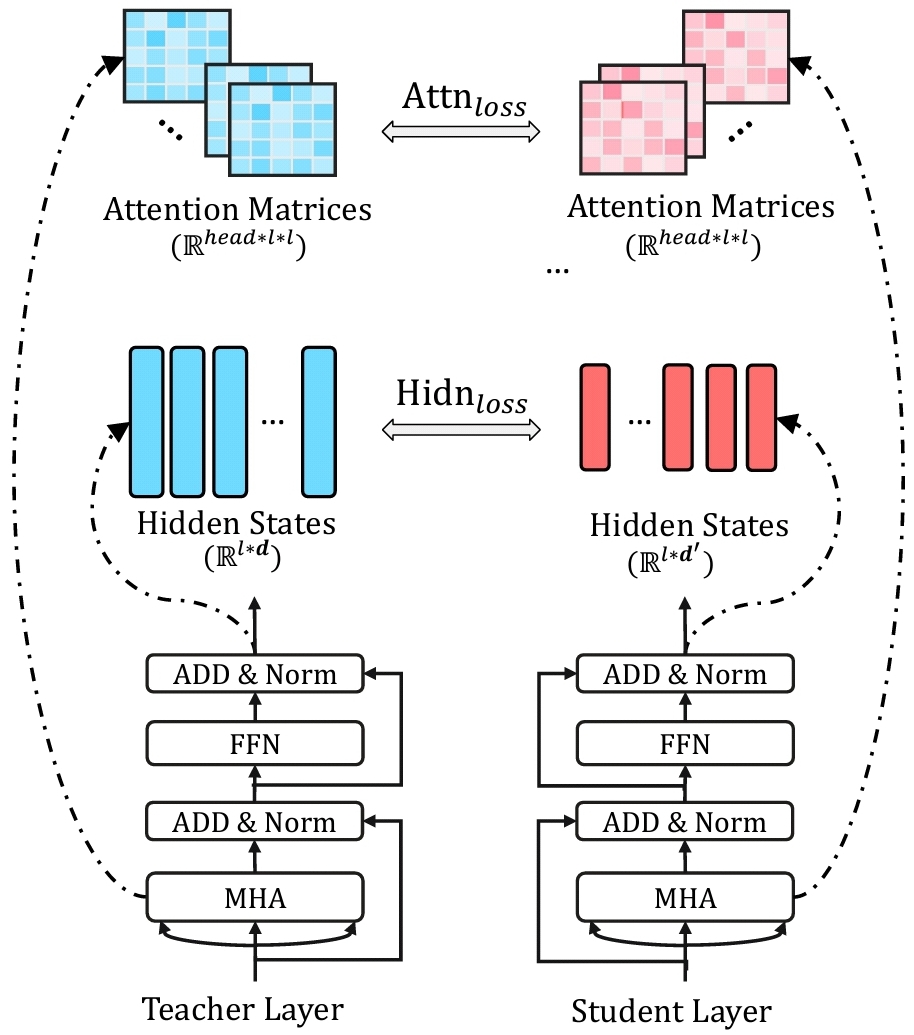}}
\caption{{\bf An overview of Transformer distillation:} (a) the framework of Transformer distillation, (b) the details of Transformer-layer distillation consisting of $Attn_{loss}$ (attention based distillation) and $Hidn_{loss}$ (hidden state based distillation)~\cite{Jiao2019}.}
\label{fig:Jiao2019}
\end{figure}
Xiaoqi Jiao {\it et al.}~\cite{Jiao2019} proposed a novel Transformer distillation method that is specially designed for knowledge distillation of the Transformer-based models. The plenty of knowledge encoded in a large “teacher” BERT can be well transferred to a small “student” TinyBERT. They also introduced a new two-stage learning framework for TinyBERT, which performs Transformer distillation at both the pre-training and task-specific learning stages. This framework ensures that TinyBERT can capture the general-domain as well as the task-specific knowledge in BERT.
Fig.~\ref{fig:Jiao2019} shows an overview of Transformer distillation.

\subsubsection{DistilBERT}
Victor Sanh {\it et al.}~\cite{Sanh2020} proposed a method to pre-train a smaller general-purpose language representation model, called DistillBERT, which can then be fine-tuned with good performances on a wide range of tasks. Here, they leverage knowledge distillation during the pre-training phase and show that it is possible to reduce the size of a BERT model by 40\%, while retaining 97\% of its language understanding capabilities and accelerating by 60\%. To leverage the inductive biases learned by larger models during pre-training, they introduced a triple loss combining language modeling, distillation, and cosine-distance losses.

\subsection{Quantization} \label{ssec:quant}
In the context of deep learning, knowledge distillation has been successfully used to compress heavy networks with a larger capacity model (teacher) to a smaller neural network (student). Quantization is also widely used to reduce parameter size and computational complexity of deep neural networks, especially for resource-constrained edge devices. 

In this paper, we review three papers~\cite{Mishra2018, Polino2018, Kim2019} to survey the cases where knowledge distillation is used for model compression via quantization. In the domain of knowledge distillation, there are a couple of distinctive approaches, based on “{\it offline knowledge distillation}” and “{\it online knowledge distillation}”. Based on this categorization (Fahad Sarfraz {\it et al.}~\cite{Sarfraz2020}), we divided the surveyed cases into the two categories:
\begin{itemize}	
\item Quantization with KD (Offline KD)
\item Quantization-aware KD (Online KD)
\end{itemize}

\subsubsection{Quantization with KD}
In the first approach, a teacher network is fixed when training the quantized student network in the teacher-student network architecture. 

Asit Mishra {\it et al.}~\cite{Mishra2018} studied the combination of quantization and KD, and showed that the performance of low-precision networks can be significantly improved by using knowledge distillation techniques. Its approach, Apprentice, achieves state-of-the-art accuracies using ternary precision and 4-bit precision for variants of ResNet architecture on ImageNet dataset. It presents three schemes on how we can apply knowledge distillation techniques to various stages of the train-and-deploy pipeline. In the first scheme, a low-precision network and a full-precision network are jointly trained from scratch using the knowledge distillation scheme. In the second scheme, it start with a full-precision trained network and transfer knowledge from this trained network continuously to train a low-precision network from scratch. In the third scheme, it starts with a trained full-precision large network and an apprentice network that has been initialized with full-precision weights. 

\begin{figure}[htbp]
\centerline{\includegraphics[width=\linewidth]{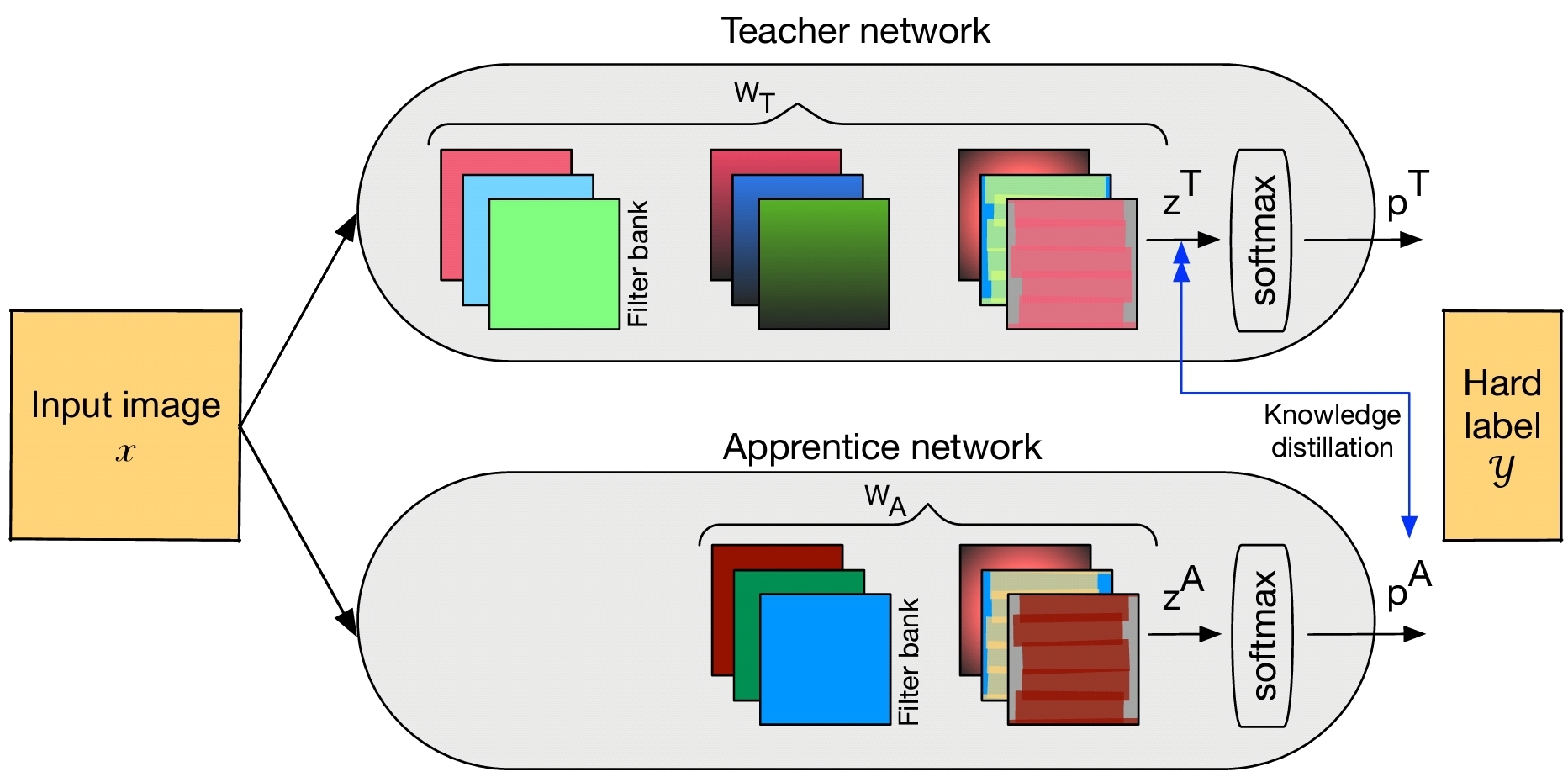}}
\caption{The schematic structure of the knowledge distillation setup~\cite{Mishra2018}.}
\label{fig:Mishra2018}
\end{figure}
The apprentice network’s precision is lowered and is fine-tuned using knowledge distillation techniques. Each of the scheme produces a low-precision model that surpasses the accuracy of the equivalent low-precision model published to date. Fig.~\ref{fig:Mishra2018} shows the knowledge distillation setup for the apprentice network. Herein, we can regard the high precision network as the teacher network and the low-precision network as the apprentice network.

\begin{algorithm}[htb]
\SetAlgoLined
\SetKwFunction{PTrain}{TrainingModelWeights}
\SetKwInput{KwInput}{Input}
\SetKwInput{KwOutput}{Output}
\SetKwProg{Fn}{Procedure}{}{}
 \KwInput{the network weights $w$, quantization level $s$} 
 \KwOutput{$w^q$}
 \While{ }{
  $w^q \gets$ quant-function$(w,s)$\;
  Run forward pass and compute distillation loss $l(w^q)$ \;
  Run backward pass and compute $\frac{\partial l(w^q)}{\partial w^q}$ \;
  Update original weights using SGD {\bf in full precision} 
  $w = w - \nu \cdot \frac{\partial l(w^q)}{\partial w^q}$
 }
 Finally quantize the weights before returning: 
 $w^q \gets$ quant-function$(w,s)$\;
\caption{Quantized Distillation}
\label{alg:QD}
\end{algorithm}

\begin{figure}[htbp]
\centerline{\includegraphics[width=\linewidth]{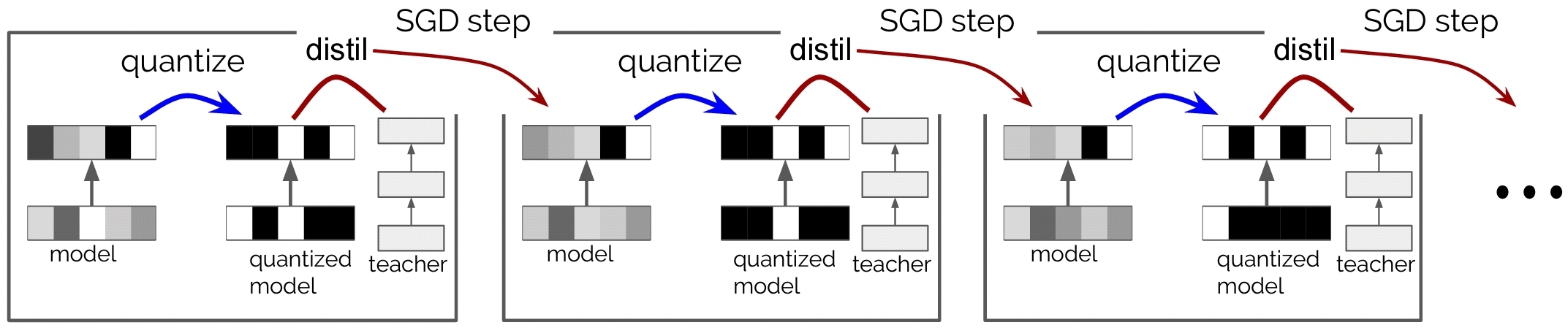}}
\caption{A depiction of the steps of quantized distillation~\cite{Polino2018}.}
\label{fig:Polino2018}
\end{figure}
Antonio Polino {\it et al.}~\cite{Polino2018} examined whether distillation and quantization can be jointly leveraged for better compression and proposed two new compression techniques (~\emph{quantized distillation} and \emph{differentiable quantization}), which jointly leverage weight quantization and distillation on larger networks, called “teacher”, into compressed “student” networks. The first method is called \emph{quantized distillation} which leverages distillation during the training process, by incorporating distillation loss, expressed with respect to the teacher network, into the training of a smaller student network whose weights are quantized to a limited set of levels. In other words, it performs the stochastic gradient descent (SGD) step on the full-precision model, but computing the gradient on the quantized model, derived with respect to the distillation loss. 
Algorithm~\ref{alg:QD} describes the quantization distillation procedure and Fig.~\ref{fig:Polino2018} shows quantization distillation steps, respectively.
Note the accumulation over multiple steps of gradients on the unquantized model leads to a switch in quantization (e.q. top layer left most square)

\begin{algorithm}[htb]
\SetAlgoLined
\SetKwFunction{PTrain}{TrainingModelWeights}
\SetKwInput{KwInput}{Input}
\SetKwInput{KwOutput}{Output}
\SetKwProg{Fn}{Procedure}{}{}
 \KwInput{the network weights $w$, the initial quantization points $p$} 
 \KwOutput{$w^q$}
 \While{ }{
  $w^q \gets$ quant-function$(w,p)$\;
  Run forward pass and compute distillation loss $l(w^q)$ \;
  Run backward pass and compute $\frac{\partial l(w^q)}{\partial w^q}$ \;
  Compute [Q(): Quantization function] $\frac{\partial Q(v,p)_i}{\partial p_j}=\begin{cases}\alpha_i, & \text{if } v_i \text{ is quantized to } p_j \\ 0, & \text{otherwise}\end{cases}$ \;
  Update original weights using SGD or similar:
  $p = p - \nu \cdot \frac{\partial l(w^q)}{\partial p}$ \;
 } 
\caption{Differentiable Quantization}
\label{alg:DQ}
\end{algorithm}
The second method, \emph{differentiable quantization}, optimizes the location of quantization points through SGD, to better fit the behavior of the teacher model. 
It is introduced as a general method of improving the accuracy of a quantized neural network, by exploiting non-uniform quantization point placement.
Algorithm~\ref{alg:DQ} shows the differentiable quantization procedure.

It shows that quantized shallow students can reach similar accuracy levels to state-of-the-art full precision teacher models, while providing up to the order of magnitude compression, and the inference speed-up factor that is almost linear to the depth reduction.

\subsubsection{Quantization-Aware KD}
The second approach adopts online knowledge distillation where a teacher network is also trained when training a quantized student network.

The inherent differences between the distributions of the full-precision teacher network and the low-precision student network may yield difficulty in knowledge transferring from teacher network to student network~\cite{Zhang2017, Mirzadeh2019}. Based on the assumption, several studies hypothesized that using a fixed teacher can limit the knowledge transfer~\cite{Kim2019}. In specific, they argued that using a fixed teacher, as in~\cite{Mishra2018, Polino2018} can limit the knowledge transfer due to the inherent differences between the distributions of the full-precision teacher model and the low-precision student network. Jangho Kim {\it et al.}~\cite{Kim2019} tackles this problem via online co-studying (CS) and offline tutoring (TU).

\begin{figure}[htbp]
\centerline{\includegraphics[width=\linewidth]{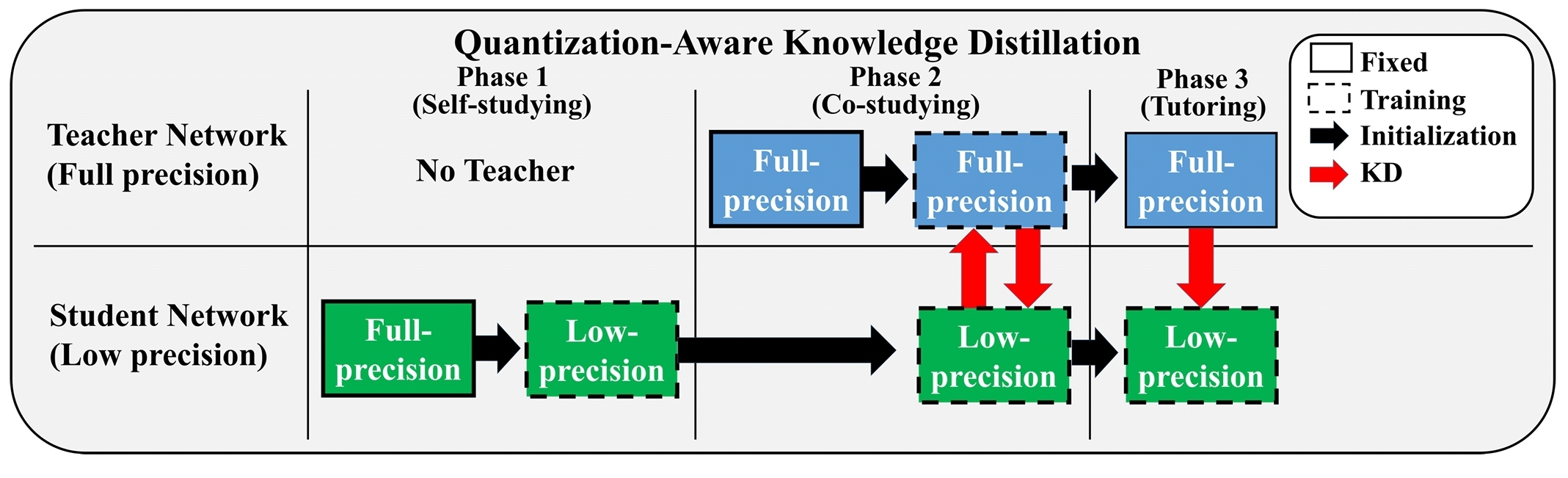}}
\caption{The overall process of QKD~\cite{Kim2019}.}
\label{fig:Kim2019}
\end{figure}
Jangho Kim {\it et al.}~\cite{Kim2019} proposed Quantization-aware Knowledge Distillation (QKD) wherein quantization and KD are carefully coordinated in three phases. First, Self-studying (SS) phase fine-tunes a quantized low-precision student network without KD to obtain a good initialization, instead of directly applying KD to the quantized student network from the beginning. Second, Co-studying (CS) phase tries to train a teacher to make it more-quantization-friendly and powerful than a fixed teacher. Finally, Tutoring (TU) phase transfers knowledge from the trained teacher to the student. This phase saves unnecessary training time and memory of the teacher network which tends to have already saturated in the co-studying phase. 
Fig.~\ref{fig:Kim2019} shows the overall process of QKD. 
Self-studying (SS) phase gives a good starting point to alleviate the low representative power and the regularization effect of KD. Co-studying (CS) phase makes a teacher model adaptable to a student model and thereby powerful than the fixed teacher. In tutoring (TU) phase, the teacher model transfers its adaptable and powerful knowledge to the student.

\section{Deep Learning Paradigm} \label{ssec:DL}

Knowledge distillation is flexibly applied to various deep learning paradigms since it was first based on the supervision of a teacher network. It is now expanding to the broad usage with weakly-supervised learning methodology and newly spotlighted unsupervised learning approach.

\subsubsection{KD in Supervised Learning}
Knowledge Distillation was originally introduced in the field of Supervised Learning~\cite{Hinton2015, Bucilua2006}. The vast majority of applications of Knowledge Distillation has been in this field to train a smaller compact network (student) under the supervision of a larger pre-trained network or an ensemble of models (teacher). In the context of deep learning, knowledge transfer has been successfully used to effectively compress the power of a larger capacity model (a teacher) to a smaller neural network (a student).

\begin{figure}[htbp]
\centering
\subfigure[\scriptsize Teacher-student Network]{
\includegraphics[width=0.3\linewidth]{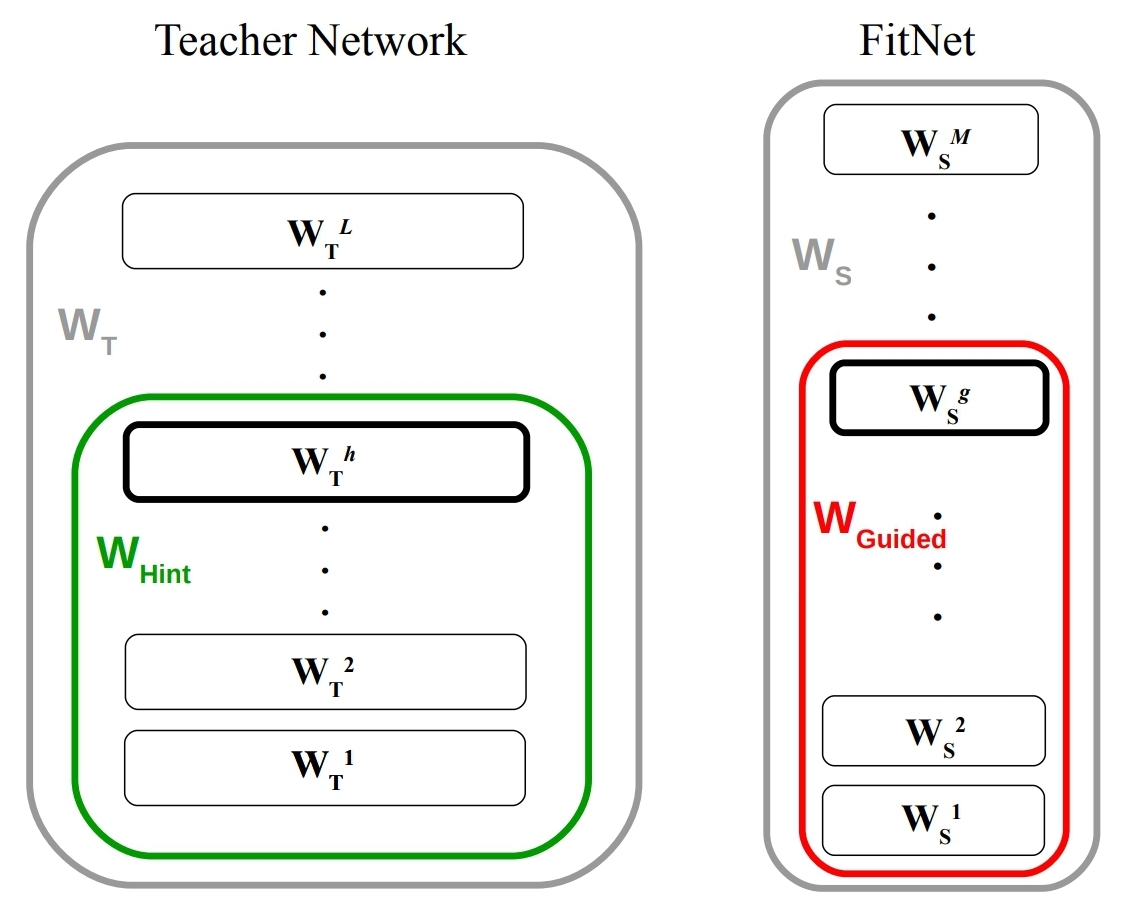}
}
\subfigure[\scriptsize Hints Training]{
\includegraphics[width=0.28\linewidth]{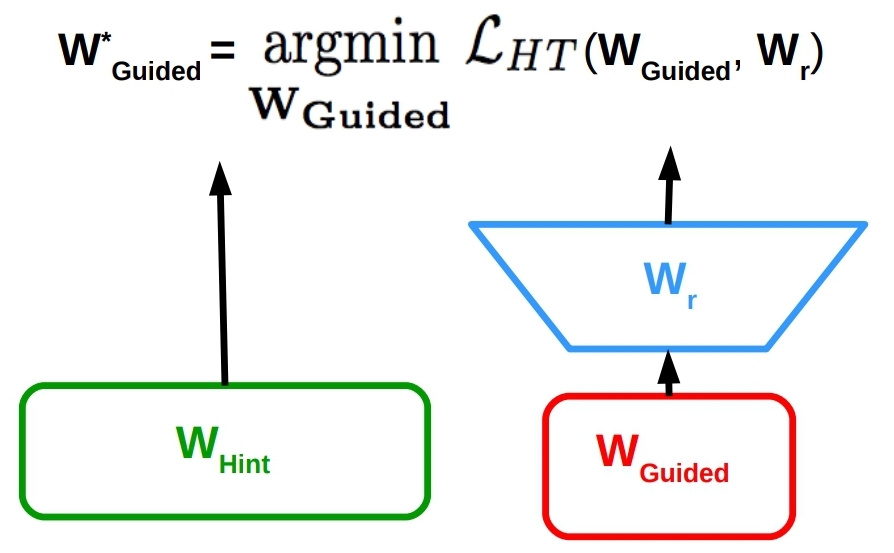}
}
\subfigure[\scriptsize Knowledge Distillation]{
\includegraphics[width=0.28\linewidth]{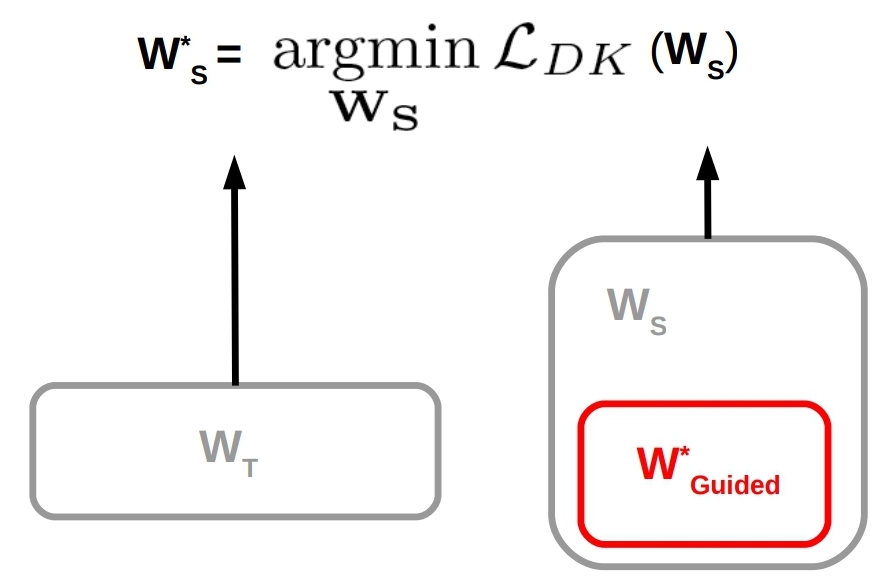}
}
\caption{Training a student network using hints~\cite{Romero2015}.}
\label{fig:Romero2015}
\end{figure}

\begin{algorithm}[htb]
\SetAlgoLined
\SetKwFunction{PTrain}{TrainingModelWeights}
\SetKwInput{KwInput}{Input}
\SetKwInput{KwOutput}{Output}
\SetKwProg{Fn}{Procedure}{}{}
 \KwInput{$\mathbf{W_S, W_T}, g, h$}
 \KwOutput{$\mathbf{W_S^{\ast}}$}
 $\mathbf{W_{Hint}} \gets \{ \mathbf{W}_{\mathbf T}^1 , \ldots, \mathbf{W}_{\mathbf T}^h \}$\; 
 $\mathbf{W_{Guided}} \gets \{ \mathbf{W}_{\mathbf S}^1 , \ldots, \mathbf{W}_{\mathbf S}^g \}$\; 
 Initialize $\mathbf{W_{r}}$ to small random values\;
 ${\mathbf W_{\mathbf{Guided}}^{\ast}} \gets {\arg\min}_{\mathbf{W_{Guided}}}  \mathcal{L}_{HT}(\mathbf{W_{Guided}}, \mathbf{W_r})$ \;
 $\{ \mathbf{W}_{\mathbf S}^1 , \ldots, \mathbf{W}_{\mathbf S}^g \} \gets \{ \mathbf{W}_{\mathbf{Guided}}^{\ast 1} , \ldots, \mathbf{W}_{\mathbf{Guided}}^{\ast g} \}$\; 
 ${\mathbf W_{\mathbf S}^{\ast}} \gets {\arg\min}_{\mathbf{W_S}}  \mathcal{L}_{KD}(\mathbf{W_S})$ \;
\caption{FitNet Stage-Wise Training}
\label{alg:FitNet}
\end{algorithm}

As an example, the work in Adriana Romero {\it et al.}~\cite{Romero2015} aimed to address the network compression problem by taking advantage of deep neural networks. It proposed an approach to train thin but deep neural networks, called FitNets, to compress wide and shallower (but still deep) networks. The method was extended to allow for thinner and deeper student models. In order to learn from the intermediate representations of teacher networks, FitNet made the student mimic the full feature maps of the teacher. However, such assumptions are too strict since the capacities of teacher and student may differ greatly. Fig.~\ref{fig:Romero2015} shows how to train a student network using hints and Algorithm~\ref{alg:FitNet} shows the FitNet training algorithm. 
The algorithm receives as input the trained parameters $\mathbf{W_T}$ of a teacher, the randomly initialized parameters $\mathbf{W_S}$ of a FitNet, and two indices $h$ and $g$ corresponding to hint/guided layers, respectively.
Let $\mathbf{W_{Hint}}$ be the teacher’s parameters up to the hint layer $h$.
Let $\mathbf{W_{Guided}}$ be the FitNet’s parameters up to the guided layer $g$. 
Let $\mathbf{W_r}$ be the regressor’s parameters. 
The first stage consists of pre-training the student network up to the guided layer, based on the prediction error of the teacher’s hint layer (line 4). 
The second stage is a KD training of the whole network (line 6).

Sergey Zagoruyko {\it et al.}~\cite{Zagoruyko2017} proposed Attention Transfer (AT) to relax the assumption of FitNet. They transferred to the student network the attention maps that summarize the full activations.

\subsubsection{KD in Weakly-Supervised Learning}
Deep learning is data-hungry. Moreover, in supervised learning, labels corresponding to input data are required to train the network. However, it is very laborious to prepare for the label sets of training data. Weakly supervised learning~\cite{Bilen2016}has been used to mitigate this problem.
Yunchao Wei {\it et al.}~\cite{YWei2018} proposed \emph{weakly supervised object detection} (WSOD) using Knowledge Distillation.  Typically, fully-supervised object detection~\cite{ShaoqingRen2015, Liu2016} requires  both labels for an object positional information (Anchor or Prior) and a class. In WSOD, only the class label is required for mining high-confidence region proposals with positive image-level annotations. The paper ~\cite{YWei2018} utilized object segmentation knowledge to take benefit of WSOD.

\begin{figure}[htbp]
\centerline{\includegraphics[width=0.99\linewidth]{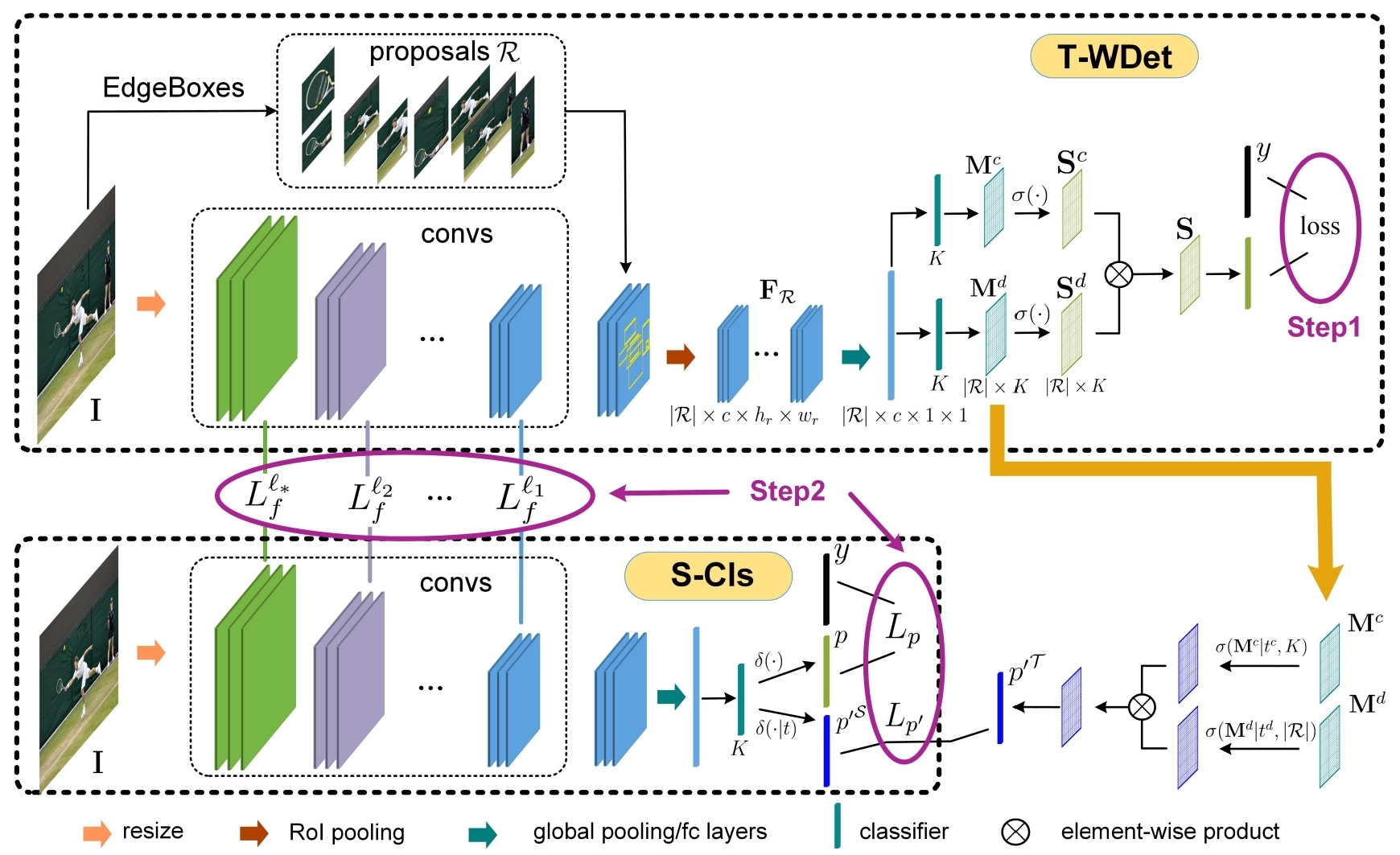}}
\caption{An overall architecture~\cite{Liu2019}}
\label{fig:Liu2019}
\end{figure}
Yongcheng Liu {\it et al.}~\cite{Liu2019} proposed a novel and efficient deep framework to boost multi-label classification by distilling knowledge from weakly-supervised detection task without bounding box annotations. This is an example of \emph{cross-task knowledge distillation}. Specifically, given the image-level annotations, (1) a weakly-supervised detection (WSD) model is developed first, and then (2) an end-to-end multi-label image classification framework is constructed which is augmented by a knowledge distillation module that guides the classification model by the WSD model according to the class-level predictions for the whole image and the object-level visual features for object RoIs. The WSD model is the \emph{teacher model} and the classification model is the \emph{student model}. 
Fig.~\ref{fig:Liu2019} shows the overall architecture of the network.
The proposed framework works with two steps: (1) we first develop a WSD model as the teacher model (called T-WDet) with only image-level annotations y; (2) then the knowledge in T-WDet is distilled into the MLIC student model (called S-Cls) via feature-level distillation from RoIs and prediction-level distillation from the whole image, where the former is conducted by optimizing the loss while the latter is conducted by optimizing the loss $L_p$ and $L'_p$.

\subsubsection{KD in Semi-Supervised Learning}

In the deep learning community, many studies agree that unsupervised learning is the future of deep learning. According to Yoshua Bengio and Yann LeCun, self-supervised learning – one of unsupervised learning approach - could lead to the creation of AI that’s more humanlike in its reasoning. Recently, knowledge distillation is finding its ways in unsupervised learning~\cite{Chen2020a,Chen2020b,Grill2020}.

\begin{figure}[htbp]
\centerline{\includegraphics[width=\linewidth]{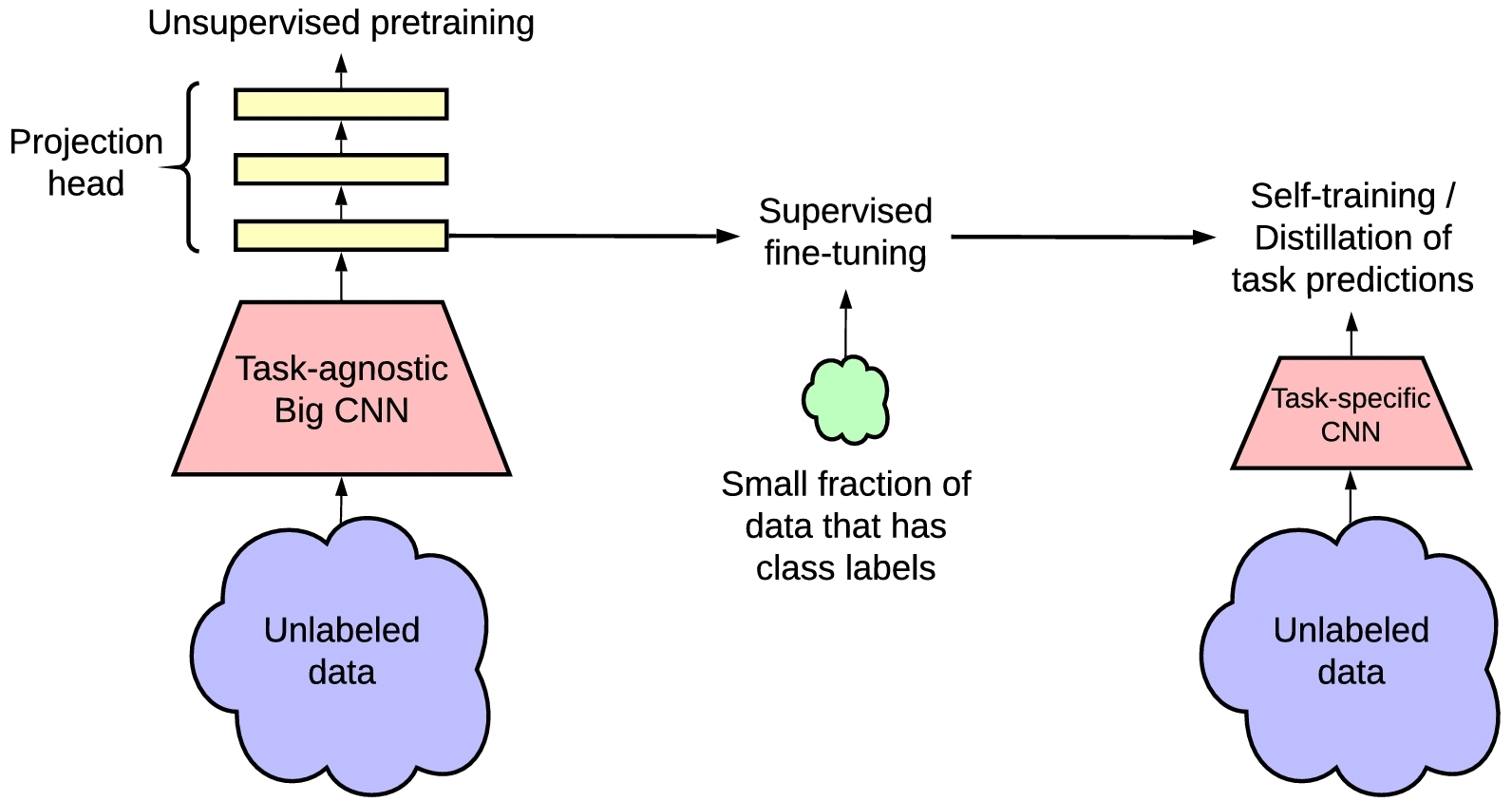}}
\caption{{\bf The proposed semi-supervised learning framework leverages unlabeled data in two ways:} (1) task-agnostic use in unsupervised pre-training, and (2) task-specific use in self-training / distillation~\cite{Chen2020a}.}
\label{fig:Chen2020}
\end{figure}
Ting Chen {\it et al.}~\cite{Chen2020a} showed that the paradigm of unsupervised pre-training followed by supervised fine-tuning is surprisingly effective for semi-supervised learning on ImageNet. The proposed semi-supervised learning algorithm can be summarized in three steps: unsupervised fine-tuning on a few labeled examples, and distillation with unlabeled examples for refining and transferring the task-specific knowledge. 
Fig.~\ref{fig:Chen2020} shows the proposed framework.

\section{Related Works} \label{sec:related}
In the perspective of machine teaching, there are a few of techniques related with knowledge distillation. Privileged information is the core concept of the smart teaching and this concept distinguishes it from knowledge distillation. David Lopez-Paz {\it et al.}~\cite{Lopez2016} tried to unify knowledge distillation and privileged information approach and analyze them analytically.

\subsection{Learning Using Privileged Information (LUPI)} \label{ssec:LUPI}
Vladimir Vapnik {\it et al.}~\cite{Vapnik2015}  incorporated an “intelligent teacher” into machine learning. 
Their solution is to consider training data formed by a collection of triplets.
\begin{align}
\{(x_1, x_1^{\ast}, y_1), \ldots, (x_n, x_n^{\ast}, y_n) \} \sim P^n(x, x^{\ast}, y),
\end{align}
where each $(x_i, y_i)$ is a feature-label pair, and the novel element $x_i^{\ast}$ is additional information about the example $(x_i, y_i)$ provided by an intelligent teacher, such as to support the learning process..

Even though, an additional information about the feature-label is provided by an intelligent teacher, the learning machine will not have an access to the teacher explanations $x_i^{\ast}$ at test time.

The framework of Learning Using Privileged Information (LUPI)~\cite{Vapnik2009, Vapnik2015} studies how to leverage the additional information $x_i^{\ast}$ at training time to build a classifier for test time that outperforms those built on the regular features alone. 
LUPI has presented a new direction in knowledge transfer by modeling the transfer of prior knowledge as a Teacher-Student interaction process. Under LUPI, a Teacher model uses Privileged Information (PI) that is only available at training time to improve the sample complexity required to train a Student learner for a given task. At a high level, PI provides some similarity information between training samples from the original feature spaces, and the Teacher hypothesis serves as additional “explanations” of the hypothesis space.

Yunpeng Chen {\it et al.}~\cite{YunpengChen2017} considered how to use PI to promote inherent diversity of a single CNN model such that the model can learn better representation and offer stronger generalization ability. To this end, they proposed a novel group orthogonal convolutional neural network (GoCNN) that learns untangled representations with in each layer by exploiting provided privileged information and enhances representation diversity effectively.

\begin{figure*}[tbp]
\centerline{\includegraphics[width=0.9\textwidth]{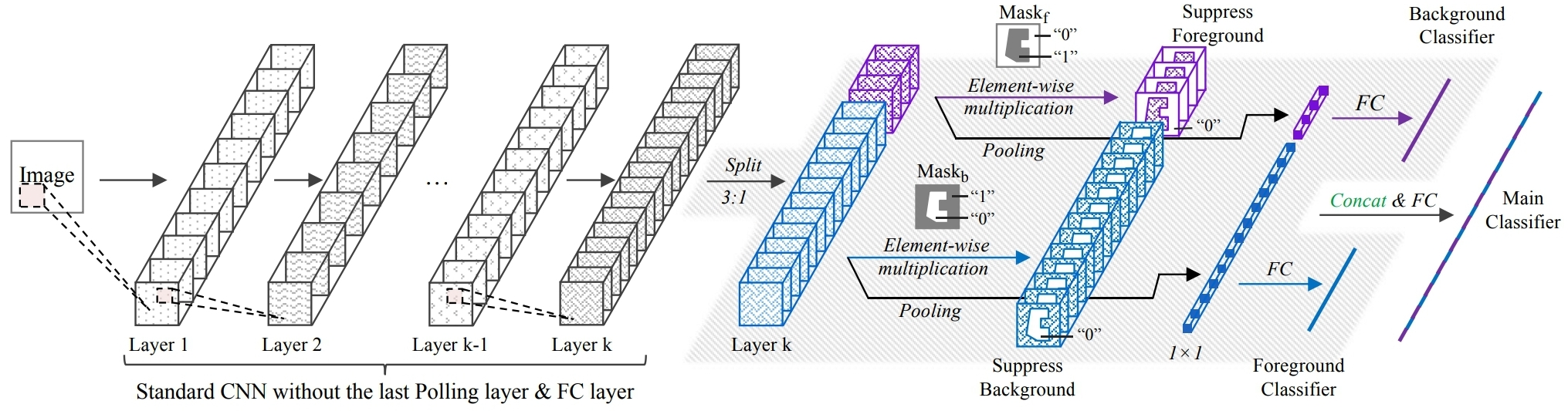}}
\caption{The architecture of GoCNN~\cite{YunpengChen2017}.}
\label{fig:YChen2017}
\end{figure*}
In this work, they proposed to exploit object segmentation annotations which are (partially) available in several public datasets as a privileged information for identifying the proper groups to give richer information. In addition, the background contents are usually independent on foreground objects within an image. 
Fig.~\ref{fig:YChen2017} shows the architecture of the GoCNN. GoCNN is built upon a standard CNN architecture where the final convolution layer are explicitly divided into two groups: the foreground group (blue) which concentrates on learning the foreground feature and the background group (purple) which learns the background feature. The output features of these two groups are concatenated as a whole representation of the input image. In testing phase, parts within the gray shadow are removed and the “Concat” (green) operation is replaced by a “Pooling” operation making GoCNN back to a standard CNN.

There are some informative papers in the theoretical and applicable aspects of the LUPI framework. D. Pechyony {\it et al.} ~\cite{Pechyony2010} gave a theoretical analysis of the LUPI framework and later M. Lapin {\it et al.}~\cite{Lapin2014} showed that LUPI is a particular instance of importance weighting.
The framework of LUPI also enjoyed multiple applications including ranking~\cite{Sharmanska2013}, computer vision~\cite{Sharmanska2014, Lopez2014}, clustering~\cite{Feyereisl2012}, metric learning~\cite{Fouad2013}, and Gaussian process classification~\cite{Hernandez2014}. 

\subsection{Generalized Distillation} \label{ssec:GD}

Knowledge distillation~\cite{Hinton2015} and privileged information~\cite{Vapnik2015} are two techniques that enable machines to learn from other machines.

There is an attempt to unify the two into generalized distillation~\cite{Lopez2016} by casting distillation as a form of learning using privileged information, a learning setting in which additional per-instance information is available at training time but not test time. 
In short, generalized distillation is a framework to learn from multiple machines and data representations. David Lopez-Paz {\it et al.}~\cite{Lopez2016} proposed the process of generalized distillation as follows and showed that generalized distillation reduces to knowledge distillation if $x_i^{\ast} = x_i$ for all $i$ with some constraints and it reduces to Vapnik’s learning using privileged information if $x_i^{\ast}$ is a privileged description of $x_i$  with some constraints.

\begin{enumerate}
\item
Learn teacher $f_t\in \mathcal{F}_t$ using the input-output pairs $\{ (x_i^{\ast}, y_i) \}_{i=1}^n$ and \eqref{eq:ft}.
\item
Compute teacher soft labels $\{\sigma(f_t(x_i^{\ast})/T)\}_{t=1}^n$, using temperature parameter $T>0$.
\item
Learn student $f_s\in \mathcal{F}_s$ using the input-output pairs $\{(x_i,y_i)\}_{i=1}^n$, $\{(x_i,s_i)\}_{i=1}^n$, \eqref{eq:fs} and imitation parameter $\lambda \in [0,1]$
\end{enumerate}

\begin{strip}
\begin{align}
f_t &= \arg \min_{f\in \mathcal{F}_t} \frac{1}{n} \sum_{i=1}^n \mathcal{L}(y_i, \sigma(f(x_i)))+ \Omega( ||f|| )  \label{eq:ft} \\
f_s &= \arg \min_{f\in \mathcal{F}_s} \frac{1}{n} \sum_{i=1}^n \left( (1-\lambda) \mathcal{L}(y_i, \sigma(f(x_i))) + \lambda \mathcal{L}(s_i, \sigma(f(x_i))) \right)  \label{eq:fs} 
\end{align}
\end{strip}
David Lopez-Paz {\it et al.}~\cite{Lopez2016} also provided theoretical and causal insight about the inner workings of generalized distillation and showed some experiments to illustrate when the distillation of privileged information is effective, and when it is not.

\section{Future Works}   \label{sec:future}

\subsection{Explainable KD} \label{ssec:explain} 

In contrast to the empirical success of knowledge distillation, there is no satisfactory theoretical explanation of this phenomenon. For example, Seyed-Iman Mirzadeh {\it et al.}~\cite{Mirzadeh2019} showed empirically the effectiveness of introducing an intermediate network between student and teacher networks. They demonstrated in the experiment that TA with any intermediate size always improves the knowledge distillation performance. 
However, one might ask \emph{“What the optimal TA size for the highest performance gain is?”}, \emph{“If one TA improves the distillation results, why not also train this TA via another distilled TA?”}. They only showed feasibility with experiments and explained the results from empirical perspectives.

To our knowledge, David Lopez-Paz {\it et al.}~\cite{Lopez2016} and  Mary Phuong {\it et al.}~\cite{Phuong2019} are the only works that examine distillation from a theoretical perspective. However, even the LUPI view conceptually falls short of explaining the effectiveness of distillation~\cite{Lopez2016}. In particular, it concentrates on the aspect that the teacher’s supervision to the student network is noise-free. Mary Phuong {\it et al.}~\cite{Phuong2019} provides the first insights into the working mechanisms of distillation by studying the special case of linear and deep linear classifiers. Specifically it proves a generalization bound that establishes fast convergence of the expected risk of a distillation-trained linear classifier.

The mathematical principles underlying distillation’s effectiveness have largely remained unexplored yet so this should be performed as a future research work.

\subsection{KD in Self-Supervised Image Representation Learning}  \label{ssec:selfLearn}

In spite of the astonishing success of supervised learning, many deep learning researchers agree that unsupervised learning is the future of deep learning because the critical shortcoming of the supervised learning is the necessity of labeled big data. This can become a big huddle when we try to apply the deep learning approach to a new field where sufficient labeled data do not exist.

In machine learning, self-supervised learning has emerged as a paradigm to learn general data representations from unlabeled examples and to fine-tune the model on labeled data. This has been particularly successful for natural language processing~\cite{Peters2018} and is an active research area for computer vision~\cite{Henaff2019}.

Recently, many studies pay attention to a variant of self-supervised learning in image understanding where the paradigm of unsupervised pre-training followed by supervised fine-tuning is surprisingly effective for semi-supervised learning on ImageNet~\cite{Chen2020b}. Knowledge distillation is actively finding its way to this promising approach as an important component in its network pipeline. Ting Chen {\it et al.}~\cite{Chen2020b} proposed the semi-supervised learning algorithm which can be summarized in three steps: unsupervised pre-tuning of a big ResNet model using SimCLRv2, supervised fine-tuning on a few labeled examples, and distillation with unlabeled examples for refining and transferring the task-specific knowledge.

\section{Conclusion}

The chronicled and comparative survey of the representative methods provide us holistic understanding and deep insight into knowledge distillation, which also motivate us to explore promising future directions.

The distillation-based approach to model compression has been proposed over a decade ago by C. Bucilu\v{a} {\it et al.}~\cite{Bucilua2006} but was re-popularized by Hinton {\it et al.}~\cite{Hinton2015}, where additional intuition about why it works – due to the additional supervision and regularization of the higher entropy soft targets – was presented.

Knowledge distillation (KD) has proven to be a promising way to induce a small model that retains the accuracy of a large model but has the smaller computational complexity. It has been working by adding a distillation loss to the usual task loss so as to encourage the student network to mimic the teacher network’s behavior. Nevertheless, a clear understanding of where valuable knowledge resides in a deep neural network is still lacking, and an optimal solution of how to capture the knowledge from a teacher network and transfer it to a student network remains an open question.


\section*{Annex}

\tableofcontents 
\listoffigures 
\listofalgorithms 

\vspace{12pt}

\end{document}